\documentclass[final,5p,times,twocolumn]{elsarticle}
\usepackage{amsmath,amssymb,amsthm,mathtools,mathrsfs,booktabs,graphicx,microtype,placeins,tikz}
\usetikzlibrary{arrows.meta,positioning}
\journal{Neurocomputing}

\usepackage{caption}
\usepackage{booktabs}
\usepackage{tabularx}
\usepackage{array}
\usepackage{float}
\usepackage{algorithm}
\usepackage{algpseudocode}
\usepackage{extarrows}
\usepackage{xcolor}
\usepackage[hidelinks]{hyperref}
\usepackage{multirow}
\newtheorem{lemma}{Lemma}
\newtheorem{theorem}{Theorem}
\newtheorem{proposition}[theorem]{Proposition}

\theoremstyle{definition}

\theoremstyle{remark}
\newtheorem{remark}[theorem]{Remark}
\newcommand{\R}{\mathbb R}
\newcommand{\E}{\mathbb E}

\newcommand{\dd}{\mathrm d}
\newcommand{\tr}{\operatorname{tr}}
\newcommand{\Null}{\operatorname{Null}}

\newcommand{\vech}{\operatorname{vech}}
\newcommand{\A}{\mathcal A}

\newcommand{\norm}[1]{\lVert#1\rVert}
\allowdisplaybreaks[2]

\providecommand{\csE}{\mathbb E}

\providecommand{\csvech}{\operatorname{vech}}

\begin{document}
\begin{frontmatter}
\title{Information-directed excitation for potential identification in generalized interacting diffusions from unlabeled data}

\author[bgi]{Chao Wang\fnref{equal}}
\author[saclay]{Zeyuan Wang\fnref{equal}}
\author[zgc]{Haiguang Liu}
\author[saclay]{Mohammed Chadli}
\author[pku]{Peijie Zhou\corref{cor1}}
\ead{pjzhou@pku.edu.cn}
\author[bgi]{Shaoshuai Li\corref{cor1}}
\ead{lishaoshuai@genomics.cn}

\address[bgi]{BGI Research, Beijing, China}
\address[saclay]{IBISC Laboratory, University Evry Paris-Saclay, Evry, France}
\address[zgc]{Zhongguancun Academy, Beijing, China}
\address[pku]{Center for Machine Learning Research, Peking University, Beijing, China}



\fntext[equal]{Equal contribution.}
\cortext[cor1]{Corresponding authors.}

\begin{abstract}

Identifying external and interaction potentials from particle observations is an inverse problem in collective dynamics.
In practice, particle correspondences across observation times are often unavailable, and passive evolution explores only part of the state space, so the potentials are hard to estimate accurately.
We address both difficulties with an active identification framework for controlled interacting diffusion.
Starting from a mobility-weighted force-error risk, 
we derive a controlled self-test loss that can be evaluated from 
unlabeled observations and known inputs, 
without numerical time derivatives. 
For linearly parametrized potentials, the risk is quadratic in the coefficients, and the loss gives joint normal equations, whose Gram matrix characterizes uniqueness and parameter distinguishability.
An error bound separates stochastic noise from observation-grid quadrature error.
Guided by this analysis, we use proximal policy optimization (PPO) to learn bounded feedback inputs, with an observable reward that balances regularized log-determinant gains against input energy.
The policy is trained in simulation with coefficients sampled from a prior, then frozen and used to collect identification data.
Experiments on a scalar diffusion with localized wells and a planar anticipated-position flock show that the learned PPO policy achieves lower coefficient estimation error with less input energy than random and sinusoidal baselines.

\end{abstract}
\end{frontmatter}

\section{Introduction}

Interacting particle systems are used to describe collective motion, aggregation, and diffusion in physical and biological populations~\cite{lu2019,lu2022}, which are usually modeled as a swarm of stochastic systems. Their evolution is influenced by external forces, mutual interactions, and stochasticity. Identifying the associated potentials from observations helps determine the forces behind the observed behavior. The accuracy of this identification depends on both the observation protocol and the configurations visited during data collection.

Parameter identification in interacting stochastic systems is complicated because each particle is subject to both an external force and interaction forces exerted by other particles.
The effects of these two types of forces must therefore be distinguished during estimation.
Several methods have been developed to account for this structure. In~\cite{lu2019}, distance-dependent interaction laws were estimated by nonparametric least squares from sampled trajectories. For stochastic systems, a regularized maximum-likelihood estimator was analyzed under a coercivity condition that controls identifiability and estimation error~\cite{lu2022}. Graph networks have also been used to infer pairwise forces and potential energies from particle accelerations, with physical laws imposed on the force aggregation~\cite{han2022}. These approaches exploit the interaction structure, but rely on trajectory or acceleration information. When the states of   particles are recorded at discrete times without labels linking individual particles across time, individual trajectories cannot be reconstructed directly.
Potential identification must then be carried out without the required trajectory information.

To address this difficulty, distribution-based and weak-form methods have been investigated. A Wasserstein-distance-based sequential Monte Carlo sampler was used for Bayesian parameter estimation from aggregate particle observations~\cite{cheng2023}. For interacting systems, interaction kernels in mean-field equations were recovered by regularized least squares from observations of the evolving distribution~\cite{lang2022}. In WSINDy, mean-field equations were combined with weak-form sparse regression to identify interaction and external potentials together with diffusivity~\cite{messenger2022}. Density estimation and importance sampling were combined in a two-phase procedure that first selects interaction terms and then refines their coefficients~\cite{shi2025}. Self-test loss functions were introduced in~\cite{gao2025}, with test functions constructed from candidate functions. For finite interacting particle systems, this approach was used to estimate external and interaction potentials through energy changes, dissipation, and diffusion~\cite{wei2026}. The resulting loss requires neither particle matching nor numerical time derivatives, and linear potential expansions give a quadratic estimation problem. In these studies, identification is performed using given observations without control input. However, how control inputs can be designed to collect more informative data is not addressed. Second, studies of interacting diffusion have often focused on one-dimensional particle models~\cite{pavliotis2022,genoncatalot2024,comte2025}. Although multidimensional states are allowed in~\cite{wei2026}, the self-test formulation is developed for uncontrolled overdamped dynamics. Controlled potential identification for generalized interacting diffusion from unlabeled observations remains less explored.
 
Outside interacting particle systems, control inputs have been designed to improve identification in several settings. Excitation trajectories were optimized for robot parameter estimation~\cite{swevers1997}, and experimental inputs were selected for cell-signaling models~\cite{bandara2009} and diffusion processes~\cite{hooker2015}. Sequential input design for neural-network identification was studied by accounting for uncertainty in the estimated parameters~\cite{hering2010}. Such designs are often guided by Fisher information, which measures the sensitivity of the observation distribution to parameter changes~\cite{ly2017}. Gram-matrix criteria are also used in regression-based system identification~\cite{manchester2010,wagenmaker2020,mania2022}. To account for the effect of current inputs on future observations, feedback policies can be learned from simulated experiments~\cite{foster2021,blau2022}. Reinforcement learning has been applied to experimental design in biology, using information-based rewards to select inputs~\cite{treloar2022}. Proximal policy optimization (PPO) provides an actor--critic method for updating continuous-input policies through a clipped surrogate objective~\cite{ppo2017}. These results motivate learned excitation, while an information criterion suited to unlabeled interacting-particle observations is still required.

Multivariate SDE inference has also been studied in other settings. Variational Bayesian inference was combined with radial basis expansions to estimate states and parameters~\cite{vrettas2010}. For related models driven by fractional noise, path signatures and attention were used to estimate parameters from observed paths~\cite{wu2026sigma}. For unpaired observations, weak collocation regression was developed to recover drift and diffusion terms from high-dimensional aggregate data through the weak Fokker--Planck equation~\cite{lu2024wcr}. Thus, high-dimensional stochastic inference is already possible under suitable model and observation assumptions. Applying these methods to finite interacting populations requires the pairwise force structure to be retained. For the present problem, both potential terms must be estimated jointly, while known   input terms must be accounted for. This motivates a self-test formulation for generalized controlled interacting diffusion.

In this paper, we investigate information-directed input design for the joint identification of external and interaction potentials from unlabeled particle observations. To formulate this identification problem in a general controlled setting, we extend the self-test loss to interacting diffusion with a known linear term in the drift, positive-definite mobility, and anisotropic or degenerate diffusion. Within this framework, prescribed basis functions for the potentials allow an accumulated Gram matrix to be computed from the observations, providing a measure of parameter distinguishability. Using this measure to guide input design, we train a bounded feedback control policy via proximal policy optimization (PPO) to collect informative samples for least-squares estimation, with the aim of improving estimation accuracy.
The main contributions of this paper are summarized as follows.
\begin{enumerate}
\item
A self-test potential estimator is developed for generalized controlled interacting diffusion using unlabeled observations. 
Equivalence between its expected loss and a weighted force-error risk is established. 
Separate basis expansions give joint least-squares equations for the external and interaction potentials without particle matching.

\item
Uniqueness of the estimate is analyzed by positive definiteness of the accumulated Gram matrix. The combination of complementary information across time windows or experiments is discussed. 
 A geometric interpretation linking a larger log determinant to a smaller volume of weakly distinguishable parameter perturbations is established.
For exact potential representations, an error bound separates martingale noise and observation-grid quadrature error and quantifies their amplification by the inverse Gram matrix.

\item
    A PPO-based active excitation method is proposed to enhance the estimation accuracy. A permutation-equivariant feedback policy uses current observations and accumulated Gram information to select bounded inputs.  Its reward is carefully designed to balance observable information gain against input energy, where the maximization of information gain is proved equivalent to minimization of the weakly distinguishable perturbations. 
    After training and validation on simulated systems with coefficients sampled
from a prior, the selected policy is fixed for new identification experiments.

\item
The method is evaluated on a scalar diffusion and a planar bird flock. Compared with uncontrolled sampling~\cite{wei2026} and the tested random and sinusoidal inputs, PPO provides more informative data and reduces both the mean and variance of estimation errors, while using less input energy than baseline inputs.
\end{enumerate}

The remainder of this paper is organized as follows. Section~\ref{sec:preliminaries} introduces some preliminaries including system model, self-test estimation, and PPO-based reinforcement learning. The controlled potential estimator and input-design method are presented in Section~\ref{sec:method}. Theoretical results are established in Section~\ref{sec:theory}. Numerical experiments are reported in Section~\ref{sec:experiments}, followed by conclusions and future work in Section~\ref{sec:discussion}.

\section{Preliminaries}\label{sec:preliminaries}
\subsection{Unlabeled particle evolution and the passive self-test loss}

We study potential identification in finite interacting particle systems using $M$ independent samples, each consisting of unlabeled observations of a fixed set of $N$ particles at $L+1$ discrete times. For simplicity, we take 
$t_\ell= \ell \Delta t$
with 
 $\Delta t=T/L$, $ \ell  = 0, \ldots, L$,
 giving 
$L+1$
observation times and
$L$
control intervals; the algorithm also accommodates nonuniform grids. Specifically, the unlabeled data are
\begin{equation}\label{eq:data}
\mathcal D=\{\mathbf X_{t_\ell}^{\pi,m}\}_{\ell=0,m=1}^{L,M},
\qquad
\mathbf X_{t_\ell}^{\pi,m}
=(X_{t_\ell}^{\pi_{t_\ell}^m(1),m},\ldots,
X_{t_\ell}^{\pi_{t_\ell}^m(N),m}),
\end{equation}
where $X_{t_\ell}^{i,m}\in\mathbb R^d$ is the state of physical particle $i$ at time $t_\ell$ in sample $m$, and $\pi_{t_\ell}^m$ is an unknown permutation of $\{1,\ldots,N\}$. Each sample records a single realization of the system, but particle correspondence across observation times is unknown. The data therefore determine only the empirical distribution $\mu_t^{N,m}=\frac1N\sum_{i=1}^N\delta_{X_t^{i,m}}$ at each observation time. Different samples have independent initial conditions and driving noises.
For the permutation-invariant energies and regression statistics below, $\mathbf X=(X^1,\ldots,X^N)$ denotes any ordered representative of an ensemble, and we suppress $m$ and $\pi$ when unambiguous.

The unknown external potential $V^\star$ depends on individual particle states, whereas the even interaction potential $\Phi^\star$ depends on relative states. Only their gradients enter the dynamics, so additive potential constants are unidentifiable. 
Consider the uncontrolled overdamped system studied in~\cite{wei2026}:
\begin{equation}\label{eq:passive-sde}
\dd X_t^i=-g_i^\star(\mathbf X_t)\dd t+\sigma\dd W_t^i,
\end{equation}
where $g_i^\star$ denotes the total potential gradient for particle $i$ under the true potentials $(V^\star,\Phi^\star)$. For candidate potentials $(V,\Phi)$, the corresponding gradient field is
\begin{equation}\label{eq:candidate-force}
g_i(\mathbf X)=\nabla V(X^i)
+\frac1N\sum_{j\ne i}\nabla\Phi(X^i-X^j),
\end{equation}
with $g_i^\star$ obtained by setting $(V,\Phi)=(V^\star,\Phi^\star)$.
Given the unlabeled data $\mathcal D$, the potential identification task is to recover $(V^\star,\Phi^\star)$ up to additive constants by solving the associated inverse problem.

The methods in~\cite{gao2025,wei2026} solve this inverse problem using the weak evolution equation for the empirical distribution, with test functions constructed from the candidate potentials and observed data. For candidate potentials $(V,\Phi)$, define the energy, dissipation, and diffusion terms by
\begin{equation}\label{eq:passive-terms}
\begin{aligned}
E_f(\mathbf X)&=\frac1N\sum_iV(X^i)
+\frac1{2N^2}\sum_{i\ne j}\Phi(X^i-X^j),\\
J_{\rm diss}(\mathbf X)&=\frac1N\sum_i\|g_i(\mathbf X)\|^2,\\
J_{\rm diff}(\mathbf X)&=\frac1N\sum_i\Delta V(X^i)
+\frac1{N^2}\sum_{i\ne j}\Delta\Phi(X^i-X^j).
\end{aligned}
\end{equation}
The corresponding loss for the uncontrolled system in~\eqref{eq:passive-sde} is
\begin{equation}\label{eq:passive-loss}
\begin{aligned}
\mathcal E_{\mathcal D}^{\rm passive}(V, \Phi)
=\frac1{MT}\sum_{m=1}^M\sum_{\ell=0}^{L-1}
\bigg[&\delta E_f^{\ell,m}
+\frac{\Delta t}{2}J_{\rm diss}(\mathbf X_{t_\ell}^{\pi,m})\\
&-\frac{\sigma^2\Delta t}{2}J_{\rm diff}(\mathbf X_{t_\ell}^{\pi,m})\bigg],
\end{aligned}
\end{equation}
where $\delta E_f^{\ell,m}
=E_f(\mathbf X_{t_{\ell+1}}^{\pi,m})
-E_f(\mathbf X_{t_\ell}^{\pi,m})$.
These quantities require neither particle pairing across times nor numerical time derivatives. Linear potential expansions make the loss quadratic in their coefficients. A computable loss alone, however, does not ensure that the observed evolution distinguishes all coefficients. The controlled estimator and informative input design below address this acquisition problem.

\subsection{Actor--critic learning and PPO}
\label{sec:ppo-background}

Reinforcement learning (RL) is a framework for learning to make sequential decisions through interaction with an environment. At each step, an agent observes the system, chooses an action, and receives a reward after the system evolves. The goal is to learn a decision rule, called a policy, that maximizes the expected cumulative reward, accounting for how current actions affect future rewards.

An actor--critic method learns two components: a policy $\pi_\omega$ (the actor) that selects actions, and a value predictor $v_\psi$ (the critic) that estimates the expected remaining reward under that policy. At decision index $\ell$, the actor selects an action $U_\ell$ from observations $O_\ell$, and then receives an immediate reward $r_\ell$. 
To account for an action's influence on both immediate and future rewards, define remaining cumulative reward, or return,
\begin{equation}\label{eq:cs-return}
R_\ell=\sum_{k=\ell}^{L-1 }
r_k,
 \quad \ell=0,\ldots,L-1.
\end{equation}
The expected remaining return under the
current policy  $\pi_\omega$ defines the value at state $s$:
\begin{equation}\label{eq:cs-values}
\mathcal V_\ell^{\pi_\omega}(s)
=\E_{\pi_\omega}[R_\ell\mid s_\ell=s].
\end{equation}
The critic $v_\psi(O_\ell)$ estimates this value from the available observation and is trained using return targets computed from recorded rewards.

The advantage measures how the expected return from a particular action $U$ differs from the policy value:
\begin{equation}\label{eq:cs-advantage}
\begin{aligned}
\mathcal A_\ell^{\pi_\omega}(s,U)
={}&\E_{\pi_\omega}[R_\ell\mid s_\ell=s,U_\ell=U] -\mathcal V_\ell^{\pi_\omega}(s).
\end{aligned}
\end{equation}
A positive advantage indicates that $U$ has a higher expected return than the policy average at that state; a negative advantage indicates a lower expected return. 
These advantages guide the Actor toward a larger expected total
return
through 
the
policy-gradient identity \cite{gae2015,ppo2017}:
\begin{equation}\label{eq:cs-gradient}
\nabla_\omega\csE_{\pi_\omega}[R_0]
=\csE_{\pi_\omega}\!\left[
\sum_{\ell=0}^{L-1}
\nabla_\omega\log\pi_\omega(U_\ell\mid O_\ell)
\,\mathcal A_\ell^{\pi_\omega}(s_\ell,U_\ell)
\right].
\end{equation}
To implement this update, advantage estimates are computed using
generalized advantage estimation \cite{gae2015}.
This method combines recorded rewards with Critic predictions,
thereby producing estimates that reflect the policy used to collect the data.
Large policy changes can make these estimates less reliable for the
updated policy, so proximal policy optimization (PPO) uses a clipped
surrogate objective to limit the additional gain from large changes
in recorded command likelihoods \cite{ppo2017}.

\section{Information-directed active identification}\label{sec:method}

 This section presents the proposed active identification method in two folds.
First, the uncontrolled overdamped
model~\eqref{eq:passive-sde} is extended to controlled stochastic
dynamics~\eqref{eq:sde}
covering overdamped systems and underdamped systems with anticipated-position coupling.
Starting from the force estimation error, we construct test functions from candidate potentials and the observed particle distribution and then use them in the weak evolution equation to derive a self-test loss computable from unlabeled observations and recorded inputs.
With linear potential expansions, minimizing this loss jointly fits
the coefficients of the external potential $V$ and interaction
potential $\Phi$.

Second, a PPO-based reinforcement learning method is employed to learn a bounded
feedback policy for informative state evolution. Its reward favors
increases in the log determinant of accumulated
force-feature information and penalizes input energy. 
After policy optimization, we freeze the policy and use it to collect $M$ informative samples, then estimate both potentials by minimizing their average self-test loss.

\begin{center}
\fbox{\parbox{0.93\linewidth}{\small
\textbf{Method overview.}\par\smallskip
\textit{Self-test formulation:}
generalized controlled dynamics
$\longrightarrow$ force estimation error
$\longrightarrow$ test-function choice
$\longrightarrow$ self-test loss
$\longrightarrow$ linear basis expansion.
\par\smallskip
\textit{Policy optimization and identification:}
train a policy with PPO
(reward: information gain $-$ input energy penalty)
$\longrightarrow$ freeze it and collect $M$ informative samples
$\longrightarrow$ minimize   self-test loss to jointly
estimate $(V,\Phi)$.}}
\end{center}

\subsection{Controlled self-test estimation}\label{sec:estimator}
\paragraph{Generalized controlled dynamics}
For $N$ homogeneous particles with states $X_t^i\in\R^d$, consider
\begin{equation}\label{eq:sde}
\dd X_t^i=[A_0X_t^i-Dg_i^\star(\mathbf X_t)+Bu_t^i]\dd t+\Sigma\dd W_t^i,
\end{equation}
where $A_0,D\in\R^{d\times d}$, $B\in\R^{d\times d_u}$, and $\Sigma\in\R^{d\times d_W}$ are known, with 
mobility
$D=D^\top\succ0$. The $W^i$ are independent standard $d_W$-dimensional Brownian motions, and $u_t^i\in\R^{d_u}$ is a known control   input. The unknown gradient force is
\begin{equation}\label{eq:force}
g_i^\star(\mathbf X)=\nabla V^\star(X^i)
+\frac1N\sum_{j\ne i}\nabla\Phi^\star(X^i-X^j),
\end{equation}
where $\Phi^\star$ is even.  Define
$ 
a_t^i=A_0X_t^i+Bu_t^i$, $C=\Sigma\Sigma^\top\succeq0$.
The particle observations ~\eqref{eq:data} are unpaired across
observation times, with inputs recorded in the same order as
the particles within each ensemble.
Each input is applied to
its assigned particle and held constant until the next observation.
In practice, the controller looks at each snapshot and assigns an input to
every particle it sees, for example by shaping a light or electric field
according to the current particle positions; the controller therefore knows
which input was given to which particle at that moment. 
By contrast, which particle is which
in the next snapshot is never needed: the policy of Section~\ref{sec:cs-ppo} uses
only a particle's own current observation and ensemble statistics, and the
estimator uses the input--particle pairing only through
$c_a(t_\ell,\mathbf X_\ell)$ in~\eqref{eq:discrete}.

For candidate potentials $(V_\theta,\Phi_\theta)$, let $g_i^\theta$ denote the gradient field
~\eqref{eq:candidate-force} of these potentials. 
The ideal identification objective is to minimize the mobility-weighted 
least-squares force risk
\begin{equation}\label{eq:risk}
\mathcal R_D(\theta)=\frac1{2NT}\E\int_0^T\sum_i
\|g_i^\theta(\mathbf X_t)-g_i^\star(\mathbf X_t)\|_D^2\dd t,
\end{equation}
Here $\E$ averages over the initial conditions and Brownian noise across samples.
To reflect how the potential gradients enter the system,
we use 
the Cameron--Martin
norm 
$\|z\|_D^2=z^\top Dz=\|-Dz\|_{D^{-1}}^2$
associated with $D$, where force error
$z$ induces drift error $-Dz$.
Expanding~\eqref{eq:risk} introduces an unknown force cross term, which the next proposition replaces in expectation with observable quantities by choosing a test function whose particle gradients equal $g_i^\theta/N$.

 
 



\begin{proposition}[Controlled self-test loss and risk equivalence]\label{prop:risk}
For   candidate potentials $(V_\theta,\Phi_\theta)$ with $\Phi_\theta$ even, define the self-test energy by
\begin{equation}\label{eq:energy}
E_{f,\theta}(\mathbf X)=\frac1N\sum_iV_\theta(X^i)
+\frac1{2N^2}\sum_{i\ne j}\Phi_\theta(X^i-X^j).
\end{equation}
Write $
\delta E_{f,\theta}=E_{f,\theta}(\mathbf X_{T})-E_{f,\theta}(\mathbf X_{0})
$ and define
$
(\mathscr D_C E_{f,\theta})(\mathbf X)
=\sum_i\tr\big(C\nabla_{X^i}^2E_{f,\theta}(\mathbf X)\big).
$
Define the continuous-observation self-test loss by
\begin{equation}\label{eq:functional-loss}
\begin{aligned}
L_T(\theta) 
={}&\frac{\delta E_{f,\theta}}{T} +\frac1T\int_0^T\bigg[
\frac1{2N}\sum_i
g_i^\theta(\mathbf X_t)^\top Dg_i^\theta(\mathbf X_t)\\
&\qquad-\frac1N\sum_i g_i^\theta(\mathbf X_t)^\top a_t^i -\frac12(\mathscr D_C E_{f,\theta})(\mathbf X_t)
\bigg]\dd t.
\end{aligned}
\end{equation}
Assume that the input $u$ is bounded, piecewise constant on the observation
intervals and adapted to the filtration generated by $(\mathbf X_0,W)$, that
\eqref{eq:sde} has a unique strong solution on $[0,T]$, and that
$
\mathbb E\!\int_0^T\sum_i\Big(\|X^i_t\|^2+\|g_i^\theta(\mathbf X_t)\|^2
+\|g_i^\star(\mathbf X_t)\|^2+\big|(\mathcal D_CE_{f,\theta})(\mathbf X_t)\big|\Big)dt<\infty $,
 $  \mathbb E\,|E_{f,\theta}(\mathbf X_t)|<\infty\ \ (t\in\{0,T\}).
 $
 Then, for each fixed $\theta$,
\begin{equation}\label{eq:risk-equivalence}
\E L_T(\theta)  =\mathcal R_D(\theta)-K_u,
\end{equation}
where
$
K_u=\frac1{2NT}\E\int_0^T\sum_i
g_i^\star(\mathbf X_t)^\top Dg_i^\star(\mathbf X_t) ~\dd t
 $
does not depend on $\theta$.
\end{proposition}
\begin{proof}
We use the permutation-invariant energy~\eqref{eq:energy} as the test function.
Evenness of the interaction potential gives
\begin{equation}\label{eq:energy-derivatives}
\begin{aligned}
\nabla_{X^i}E_{f,\theta}&=g_i^\theta/N,\\
\nabla_{X^i}^2E_{f,\theta}
&=\nabla^2V_\theta(X^i)/N
+\frac1{N^2}\sum_{j\ne i}\nabla^2\Phi_\theta(X^i-X^j).
\end{aligned}
\end{equation}
It\^o's formula  therefore yields
\begin{equation}\label{eq:ito}
\begin{aligned}
\dd E_{f,\theta}
=\bigg[&-\frac1N\sum_i
(g_i^\theta)^\top Dg_i^\star
+\frac1N\sum_i(g_i^\theta)^\top a_t^i +\frac12\mathscr D_C E_{f,\theta}\bigg]\dd t
+\dd M_t^\theta,
\end{aligned}
\end{equation}
Integrating~\eqref{eq:ito} over $[0,T]$ and rearranging gives
an identity for
the time integral of the force cross term
\begin{equation}
\begin{aligned}
&\frac1{NT}\int_0^T\sum_i
g_i^\theta(\mathbf X_t)^\top Dg_i^\star(\mathbf X_t)\dd t =-\frac{
\delta E_{f,\theta} 
}{T} 
\\
& \quad
+\frac1{NT}\int_0^T\sum_i
g_i^\theta(\mathbf X_t)^\top a_t^i\dd t 
+\frac1{2T}\int_0^T
(\mathscr D_C E_{f,\theta})(\mathbf X_t)\dd t
+\frac{M_T^\theta}{T}.
\end{aligned}
\label{eq:integrated-cross}
\end{equation}
For fixed $\theta$, the integrability assumption gives
$\mathbb E\int_0^T\|\Sigma^\top g_i^\theta(\mathbf X_t)\|^2dt
\le\|\Sigma\|^2\,\mathbb E\int_0^T\|g_i^\theta(\mathbf X_t)\|^2dt<\infty$, so
$M^\theta$ is a square-integrable martingale and $\mathbb EM_T^\theta=0$; the
remaining terms of~\eqref{eq:integrated-cross} are integrable, so expectations can be
taken term by term.
Substituting the resulting identity for the expected force cross term
into the expansion of~\eqref{eq:risk} gives
$\mathcal R_D(\theta)=\E
L_T(\theta) +K_u$,
where $L_T(\theta)$ is the self-test loss defined
in~\eqref{eq:functional-loss}.
Rearranging yields~\eqref{eq:risk-equivalence}.


\end{proof}

\paragraph{Basis-function expansions}

For identifiability analysis, we restrict the potentials to finite-dimensional linear spaces~\cite{lu2022,wei2026}. Specifically, let $\{\psi_k\}_{k=1}^{K_V}$ and $\{\varphi_s^{\rm sym}\}_{s=1}^{K_\Phi}$ be separate, linearly independent $C^2$ families, fixed before estimation, with each $\varphi_s^{\rm sym}$ even. We write
\begin{equation}\label{eq:dictionaries}
V_\alpha=\sum_{k=1}^{K_V}\alpha_k\psi_k,
\qquad \Phi_\beta=\sum_{s=1}^{K_\Phi}\beta_s\varphi_s^{\rm sym}.
\end{equation}
Let $\theta=(\alpha^\top,\beta^\top)^\top\in\R^K$, $K=K_V+K_\Phi$.  
Under this linear parameterization, the self-test loss in Proposition~\ref{prop:risk} is quadratic in the parameter vector $\theta\in\R^K$. To derive the corresponding normal equations, we first express the energy, candidate force, and diffusion term as linear functions of $\theta$:


\begin{equation}\label{eq:energy-vectors}
E_{f,\theta}=\theta^\top h,\quad
g_i^\theta=F_i\theta,\quad
\mathscr D_C E_{f,\theta}=\theta^\top\delta_C,
\end{equation}
where 
$h\in\mathbb R^K$ 
collects the prescribed energy
basis functions:
\begin{equation}
h_{V,k}=\frac1N\sum_i\psi_k(X^i),
\quad
h_{\Phi,s}=\frac1{2N^2}\sum_{i\ne j}\varphi_s^{\rm sym}(X^i-X^j),
\end{equation}
then the force basis $F_i\in\mathbb R^{d\times K}$
and diffusion basis $\delta_C\in\mathbb R^K$
 follow by differentiation:
\begin{equation}
\label{eq:diffusion-basis}
(F_i)_{\cdot r}=N\nabla_{X^i}h_r,\quad
(\delta_C)_r=\sum_{j=1}^N\operatorname{tr}
\big(C\nabla_{X^j}^2h_r\big),
\quad 1\le r\le K.
\end{equation}
The factor $N$ in $F_i$ follows from
$\nabla_{X^i}E_{f,\theta}=g_i^\theta/N$
in Proposition~\ref{prop:risk}.

The remaining terms in~\eqref{eq:functional-loss} are
$\frac1{2N}\sum_i(g_i^\theta)^\top Dg_i^\theta$
and $-\frac1N\sum_i(g_i^\theta)^\top a_t^i$.
To write these two terms as
$\frac12\theta^\top G_D\theta$
and $-\theta^\top c_a$, respectively, define  
\begin{equation}\label{eq:gram}
G_D(\mathbf X)=\frac1N\sum_iF_i^\top DF_i,\qquad
c_a(t,\mathbf X)=\frac1N\sum_iF_i^\top a_t^i.
\end{equation}
Combining these expressions with the 
candidate energy and diffusion
identities in~\eqref{eq:energy-vectors} gives the following
quadratic form.

\begin{lemma}[Quadratic form of the self-test loss]
\label{lem:normal-equations}
For a fixed observed sample and recorded input, the self-test loss in
Proposition~\ref{prop:risk} satisfies
\begin{equation}\label{eq:quadratic}
L_T(\theta) 
=\tfrac12\theta^\top A_T\theta-b_T^\top\theta,
\end{equation}
where
\begin{equation}\label{eq:normal-components}
\begin{aligned}
A_T&=\frac1T\int_0^T G_D(\mathbf X_t)\dd t,\\
b_T&=\frac{h(\mathbf X_{0})-h(\mathbf X_{T})}{T}+\frac1T\int_0^T
\left[c_a(t,\mathbf X_t)+\frac12\delta_C(\mathbf X_t)\right]\dd t.
\end{aligned}
\end{equation}
\end{lemma}

\begin{proof}
Substituting~\eqref{eq:energy-vectors} and~\eqref{eq:gram}
into~\eqref{eq:functional-loss} gives
\[
\begin{aligned}
L_T(\theta)={}&\frac{\theta^\top[h(\mathbf X_T)-h(\mathbf X_0)]}{T} 
\\
&+\frac1{2T}\int_0^T
\theta^\top G_D(\mathbf X_t)\theta\,\dd t\\
&-\frac1T\int_0^T\theta^\top
\left[c_a(t,\mathbf X_t)+\frac12\delta_C(\mathbf X_t)\right]\dd t.
\end{aligned}
\]
Since $\theta$ is constant in time, it can be taken outside
the integrals. By~\eqref{eq:normal-components}, the quadratic
integral equals $\theta^\top A_T\theta/2$, while the endpoint
term together with the last integral equals
$-\theta^\top b_T=-b_T^\top\theta$.
This proves~\eqref{eq:quadratic}.
\end{proof}


Because $D=D^\top\succ0$, both $G_D(\mathbf X)$ and $A_T$ are symmetric positive semidefinite. Differentiating~\eqref{eq:quadratic} yields the normal equations
\begin{equation}\label{eq:normal}
\nabla_\theta L_T(\theta)=A_T\theta-b_T,
\qquad A_T\widehat\theta_T=b_T.
\end{equation}
If $A_T\succ0$, the estimator is
\begin{equation}\label{eq:optimal-estimate}
\widehat\theta_T=\operatorname*{arg\,min}_{\theta\in\R^K}L_T(\theta)
=A_T^{-1}b_T.
\end{equation}

The input $u$ enters $b_T$ through $a_t^i$ and affects both
$A_T$ and $b_T$ through the observed evolution.
Since $g_i^\theta$ contains both external and interaction forces,
expanding $(g_i^\theta)^\top Dg_i^\theta$ produces cross terms
between their basis forces. This generally leads to nonzero mixed
blocks in $A_T$, so the normal equations couple the two groups
of coefficients, which are estimated jointly.

\paragraph{Estimator from $M$ samples}
To estimate the loss from discrete observations, we approximate the integrals
in~\eqref{eq:normal-components} by one-sided Riemann sums on the grid
$t_\ell=\ell\Delta t$, $\ell=0,\dots,L$, with $\Delta t=T/L$, and average over
$M$ independent samples.
Write $\mathbf X_\ell^m=\mathbf X_{t_\ell}^{\pi,m}$ for the $m$-th observed
ensemble at time $t_\ell$.
The inputs applied on $[t_\ell,t_{\ell+1})$, $\ell=0,\dots,L-1$, are recorded in
the same order as the particles observed at $t_\ell$, so that
$c_a(t_\ell,\mathbf X_\ell^m)$ in~\eqref{eq:gram} can be evaluated from the
recorded ensemble and inputs.
The Gram matrix and the right-hand side use opposite ends of each control
interval: $G_D$ is evaluated at the ensemble $\mathbf X_{\ell+1}^m$ produced by
the interval, so that the Gram matrix collects exactly the ensembles generated
by the applied inputs and coincides with the information accumulator used for
the reward in Section~\ref{sec:cs-ppo}; the integrands of
$b_T$ are evaluated at the start $\mathbf X_\ell^m$ of the interval, where the
pairing between particles and inputs is recorded.
Both are first-order quadrature rules with $L$ nodes of weight
$\Delta t/T=1/L$.
The averaged coefficients are
\begin{equation}\label{eq:discrete}
\begin{aligned}
A_{M,\Delta t}
&=\frac{\Delta t}{MT}\sum_{m=1}^M\sum_{\ell=1}^{L}
G_D(\mathbf X_\ell^m),\\
b_{M,\Delta t}
&=\frac1{MT}\sum_{m=1}^M
\bigl[h(\mathbf X_0^m)-h(\mathbf X_L^m)\bigr]
\\
&\quad
+\frac{\Delta t}{MT}\sum_{m=1}^M\sum_{\ell=0}^{L-1}
\left[c_a(t_\ell,\mathbf X_\ell^m)+\frac12\delta_C(\mathbf X_\ell^m)\right].
\end{aligned}
\end{equation}
For a single sample ($M=1$), we denote the corresponding coefficients by
$A_{1,\Delta t}^{m}$ and $b_{1,\Delta t}^{m}$. The empirical loss is
\[
L_{M,\Delta t}(\theta)
=\tfrac12\theta^\top A_{M,\Delta t}\theta-b_{M,\Delta t}^\top\theta.
\]
If $A_{M,\Delta t}\succ0$, its unique minimizer is
$\widehat\theta_{M,\Delta t}=A_{M,\Delta t}^{-1}b_{M,\Delta t}$.
No particle pairing across observation times is required: every term
of~\eqref{eq:discrete} uses a single observed ensemble together with the inputs
recorded at that same time.
\subsection{PPO-based informative excitation}\label{sec:cs-ppo}
The Gram matrix $A_T$ in~\eqref{eq:normal-components} depends on the
particle ensembles visited during the experiment.
Under passive sampling, different coefficient vectors may produce
similar forces on these ensembles, making them difficult to
distinguish from the data.
We therefore design bounded inputs that modify the particle evolution
to improve the accumulated force-feature information.
To account for the effect of inputs on future observations, 
we formulate input selection as a finite-horizon reinforcement learning
problem, with rewards that balance information gain against input
energy, and learn a feedback policy using the actor--critic and PPO
framework of Section~\ref{sec:ppo-background}.


\paragraph{Accumulated information and reward}
Each sample evolves according to~\eqref{eq:sde}.
At $t_\ell$ ($\ell = 0, \ldots, L-1$), the input $U_\ell=(u_\ell^i)_{i=1}^N$ is held on
$[t_\ell,t_{\ell+1})$, forming one population-level RL transition.
To measure the information obtained after applying this input, we
update the accumulated force-feature Gram matrix using the next
observed ensemble:
\begin{equation}\label{eq:cs-H}
H_0 = 0 , 
~~~
H_{\ell+1} =H_\ell+\frac{\Delta t}{T}
G_D(\mathbf X_{t_{\ell+1}}^\pi)
,
 ~~~ \ell=0,\ldots,L-1.
\end{equation} 

Consequently, $H_\ell=\frac{\Delta t}{T}\sum_{k=1}^{\ell}G_D(X_{t_{k}}^\pi)$,
and the final accumulated matrix is the estimator Gramian 
$H_{L }=A_{1,\Delta t}$ within a single sample.
We use the regularized log determinant to score accumulated information: 
\begin{equation}\label{eq:cs-score}
J(H)=\frac1K\log\det(H+\varepsilon_GI_K),
\qquad \varepsilon_G>0,
\end{equation}
where the fixed regularization $\varepsilon_G>0$ ensures positive
definiteness, and 
a parameter-volume interpretation of this score is given in
Section~\ref{sec:volume}.
We reward the score increment and penalize input energy:
\begin{equation}\label{eq:cs-reward}
r_\ell=J(H_{\ell+1})-J(H_\ell)-\beta_u e_\ell,
\quad
e_\ell=
\frac{\Delta t}{NT}
\sum_i\|u_\ell^i\|^2 
\end{equation}
where 
$\beta_u\ge0$ weights input energy.
For the undiscounted return, the information increments telescope:
\begin{equation}\label{eq:cs-telescope}
\sum_{\ell=0}^{L-1}r_\ell
=J(H_L)-J(H_0)-\beta_u E_u,
\qquad E_u=\sum_{\ell=0}^{L-1}e_\ell.
\end{equation}
Since $J\left(H_0\right)=\log \varepsilon_G$ is a constant, 
maximizing expected return is equivalent to maximizing $\mathbb{E}_{\pi_\omega}\left[J\left(A_{1, \Delta t}\right)-\beta_u E_u\right]$. 
Reward evaluation requires only known force basis functions,
observed ensembles, and recorded inputs, without particle pairing
across time.
Section~\ref{sec:theory} analyzes how this information objective
relates to estimation accuracy.

\paragraph{Permutation-equivariant actor and invariant critic}
At each decision time $t_\ell$, 
$\ell = 0, \ldots, L-1$, 
the observation for particle $i$ is
\begin{equation}\label{eq:cs-observation}
o_\ell^i=[(X_{t_\ell}^i)^\top,\bar X_\ell^\top,
\vech(S_\ell)^\top,\vech(H_\ell)^\top,t_\ell/T]^\top,
\end{equation}
where the ensemble mean and covariance are
\[
\bar X_\ell=\frac1N\sum_iX_{t_\ell}^i,\quad
S_\ell=\frac1N\sum_i(X_{t_\ell}^i-\bar X_\ell)
(X_{t_\ell}^i-\bar X_\ell)^\top.
\]
and $\csvech(H_\ell)$ stacks the
lower-triangular entries of $H_\ell$ row by row.
Denoting $O_\ell=(o_\ell^i)_{i=1}^N$, the Actor selects a population
command through the policy
\begin{equation}\label{eq:cs-policy}
U_\ell\sim\pi_\omega(\cdot\mid O_\ell),
\end{equation}
where $\omega$ contains the trainable Actor parameters.
All particles share the same Actor network, with mean output
$\mu_\ell^i=\mu_\omega(\widetilde o_\ell^i)$;
$\widetilde o_\ell^i$ denotes the standardized and clipped observation.
During training, commands are sampled independently across particles
conditional on $O_\ell$:
\begin{equation}\label{eq:cs-action}
z_\ell^i\sim\mathcal N(\mu_\ell^i,e^{2\nu}),
\qquad
u_\ell^i=u_{\max}\tanh z_\ell^i,
\end{equation}
where $\nu$ is a shared, learned log standard deviation included in
$\omega$. Each command is applied to its observed particle for the hold
interval.

The Critic is trained to approximate this conditional expectation with
$v_\psi(O_\ell)$, using return targets from the recorded rewards.
It predicts the value from averaged particle features:
\begin{equation}\label{eq:cs-critic}
v_\psi(O_\ell)=h_\psi\!\left(\frac1N\sum_i g_\psi(o_\ell^i)\right),
\end{equation}
where $g_\psi$ is a shared feature map and $h_\psi$ maps the average to a scalar, making the prediction invariant to particle permutations.
We train the actor and critic using PPO  on rollouts of~\eqref{eq:sde}
whose coefficients are sampled from a prior $p(\theta)$.
After training, we select the policy checkpoint with the highest mean
validation return validated under the same prior. Using the
deterministic inputs
\begin{equation}\label{eq:cs-frozen}
u_\ell^i=u_{\max}\tanh\mu_\omega(\widetilde o_\ell^i),
\end{equation}
we collect $M$ new independent samples  from the true unknown system and
estimate the potentials through self-test loss.


\begin{algorithm}[t]
\caption{Information-directed active identification}
\label{alg:active}
\begin{algorithmic}[1]
\Require $A_0,D,B,\Sigma$ and potential bases; prior $p(\theta)$; $N,T,L,M,u_{\max}$.
\Ensure Estimated potentials $V_{\widehat\alpha}$ and $\Phi_{\widehat\beta}$.

\Statex \textbf{Train the policy}
\State Initialize the shared actor and pooled critic.
\For{each PPO update}
           \State Collect independent rollouts of~\eqref{eq:sde} with coefficients
       $\tilde\theta\sim p(\theta)$, inputs~\eqref{eq:cs-action}
           and rewards~\eqref{eq:cs-reward}.
    \State Update the actor and critic using PPO.
\EndFor
\State Select the checkpoint with highest mean validation return;
       freeze the actor and observation normalization.

\Statex \textbf{Collect identification data}
\State Collect $M$ new independent samples from the true unknown system
       using~\eqref{eq:cs-frozen}; retain observed ensembles and applied inputs.
\Statex \textbf{Estimate the potentials}
\State Form $A_{M,\Delta t}$ and $b_{M,\Delta t}$
       using~\eqref{eq:discrete}.
\State For $A_{M,\Delta t}\succ0$,
       solve $A_{M,\Delta t}\widehat\theta=b_{M,\Delta t}$.
\State \Return $V_{\widehat\alpha}$ and $\Phi_{\widehat\beta}$.
\end{algorithmic}
\end{algorithm}



\section{Theoretical analysis}\label{sec:theory}
The method of Section~\ref{sec:method} raises three questions: when the observed evolution uniquely determines the coefficients; how the log determinant quantifies parameter distinguishability; and how stochasticity and observation-grid errors affect estimation. We address these questions in turn.
Section~\ref{sec:identifiability} characterizes uniqueness through positive definiteness of the accumulated Gram matrix.
Section~\ref{sec:volume} shows that, for a fixed response threshold and parameterization, maximizing its log determinant minimizes the volume of weakly distinguishable parameter perturbations.
Section~\ref{sec:error} derives error bounds that separate stochasticity and observation-grid quadrature residuals and quantify their amplification by the inverse Gram matrix.
 
\subsection{Uniqueness and aggregate excitation}\label{sec:identifiability}
For a parameter perturbation $v\in\R^K$,
the mean $D$-weighted squared force change induced by
that perturbation on the visited particle ensembles is
\begin{equation}\label{eq:gram-quadratic}
v^\top A_Tv=\frac1{NT}\int_0^T\sum_i
\norm{D^{1/2}F_i(\mathbf X_t)v}^2\dd t.
\end{equation}
A zero value indicates that two candidate parameters are indistinguishable in the experiment.
The following proposition relates uniqueness to the absence of
such undetectable directions. 

\begin{proposition}[Uniqueness and aggregate excitation]
\label{prop:unique}
The unrestricted loss \eqref{eq:quadratic} has a unique finite
minimizer if and only if $A_T\succ0$.
For any finite collection of symmetric positive semidefinite
window or experiment Gram matrices $A_j$ and positive weights $w_j$,
\begin{equation}\label{eq:null-sum}
\Null\Big(\sum_jw_jA_j\Big)=\bigcap_j\Null(A_j),
\end{equation}
where $\Null(Q):=\{v\in\R^K:Qv=0\}$ denotes the nullspace
of $Q\in\R^{K\times K}$.
\end{proposition}

\begin{proof}
By \eqref{eq:gram-quadratic}, $A_T\succeq0$.
If $A_T\succ0$, set $\widehat\theta_T=A_T^{-1}b_T$.
Expanding \eqref{eq:quadratic} gives, for every $v\in\R^K$,
\begin{equation}
L_T(\widehat\theta_T+v)-L_T(\widehat\theta_T)
=\tfrac12v^\top A_Tv.
\label{eq:vav_proof}
\end{equation}
This is strictly positive for $v\ne0$, so $\widehat\theta_T$ is
the unique finite minimizer.
Conversely, suppose a unique minimizer $\widehat\theta_T$ exists.
It satisfies \eqref{eq:normal}, so the same identity \eqref{eq:vav_proof} holds.
If $A_T$ were singular, a nonzero
$v\in\Null(A_T)$ would give another minimizer, contradicting uniqueness.

For the aggregation identity, positive semidefiniteness gives
\[
v\in\Null\Big(\sum_jw_jA_j\Big)
\quad\Longleftrightarrow\quad
\sum_jw_j\norm{A_j^{1/2}v}^2=0.
\]
Since $w_j>0$, this holds exactly when $A_j^{1/2}v=0$,
equivalently $A_jv=0$, for every $j$, proving \eqref{eq:null-sum}.
\end{proof}

The condition does not require every instant or every window to identify every coefficient. Complementary singular windows can be aggregated into positive-definite information. Thus finite-horizon identifiability can follow from complementary information across windows without requiring each window to be positive definite.

\begin{remark}
	Denote  $A_T$ as $A_T = \begin{bmatrix}A_{VV}&A_{V\Phi}\\A_{\Phi V}&A_{\Phi\Phi}\end{bmatrix}$.
	For $A_{VV}\succ0$, the Schur complement condition is $A_{\Phi\Phi}-A_{\Phi V}A_{VV}^{-1}A_{V\Phi}\succ0$. Thus uniqueness depends jointly on the diagonal information blocks and the correlations between external and interaction force features.
\end{remark}

The input affects identifiability by changing the evolution of the
particle ensemble and hence the time-averaged Gram matrix $A_T$.
When passive trajectories provide little information about a parameter direction, a control input can help by driving the system to particle ensembles where that direction produces a distinguishable force change. Information from different time windows can then be combined according to \eqref{eq:null-sum}.
If passive data already distinguish all parameter directions, the purpose of control is to increase information in weakly excited directions within the available observation time. Conversely, a nonzero parameter direction $v$ satisfying $F_i(\mathbf X)v=0$ for every    particle ensemble cannot be identified by any input. 

\subsection{Parameter volume and the log determinant}\label{sec:volume}

Proposition~\ref{prop:unique} characterizes finite-time uniqueness by
positive definiteness of the Gram matrix. This excludes exact confounding,
but does not
quantify how strongly different parameter values are distinguished by the
observed particle ensembles. 
Even when the minimizer is unique, parameter perturbations along certain
directions may induce only small force changes, making the
corresponding parameter variations difficult to detect.
We quantify how well parameter values are distinguished by measuring
the volume of perturbations whose mean weighted squared force change
does not exceed a prescribed threshold.

\begin{proposition}[Volume of weakly distinguishable perturbations]
\label{prop:identifiability-volume}
Assume $A_T\succ0$.
For a response threshold $c>0$, define
\begin{equation*}
\mathcal U_c(A_T)
:=\bigl\{v\in\mathbb R^K:v^\top A_T v\le c\bigr\}.
\end{equation*}
This set is an ellipsoid with finite volume
\begin{equation}\label{eq:cs-volume}
\operatorname{Vol}\bigl(\mathcal U_c(A_T)\bigr)
=\omega_K c^{K/2}\det(A_T)^{-1/2},
\end{equation}
where $\omega_K$ is the volume of the unit ball in $\mathbb R^K$.
\end{proposition}

\begin{proof}
Let $\mathbb B_K:=\{z\in\R^K:\norm{z}\le1\}$.
Since $A_T\succ0$, the invertible linear change of variables
$v=c^{1/2}A_T^{-1/2}z$ gives
$v^\top A_Tv=c\norm{z}^2$. Hence
\[
\mathcal U_c(A_T)=c^{1/2}A_T^{-1/2}\mathbb B_K,
\]
so $\mathcal U_c(A_T)$ is an ellipsoid.
To compute its volume, 
since 
by the change-of-variables formula, 
the absolute Jacobian determinant is
$ 
\left|\det\left(\frac{\partial v}{\partial z}\right)\right|=\det\bigl(c^{1/2}A_T^{-1/2}\bigr)
=c^{K/2}\det(A_T)^{-1/2} 
$, 
injecting it into $\operatorname{Vol}\bigl(\mathcal U_c(A_T)\bigr)$ therefore   yields
\begin{equation}
\operatorname{Vol}\bigl(\mathcal U_c(A_T)\bigr)
=\int_{\mathbb B_K}
\left|\det\left(\frac{\partial v}{\partial z}\right)\right|
\,\mathrm dz=\omega_K c^{K/2}\det(A_T)^{-1/2}<\infty.
\end{equation}
\end{proof}

\begin{remark}
(1)
Taking logarithms in \eqref{eq:cs-volume} gives
\begin{equation*}
\log\operatorname{Vol}\bigl(\mathcal U_c(A_T)\bigr)
=\log\omega_K+\frac K2\log c
-\frac12\log\det(A_T).
\end{equation*}
Thus, for fixed $c$ and fixed parameter coordinates, maximizing
$\log\det(A_T)$ over positive-definite Gram matrices is equivalent
to minimizing the volume of weakly distinguishable parameter
perturbations.

(2)
The shape of $\mathcal U_c(A_T)$ describes how the response
threshold $c$ constrains parameter perturbations in different
directions. 
Let $e_1,\ldots,e_K$ be an orthonormal eigenbasis of $A_T$,
with corresponding eigenvalues $\lambda_k>0$.
For a perturbation $v=a e_k$, we have
$v^\top A_Tv=\lambda_k a^2$, and hence
\[
v\in\mathcal U_c(A_T)
\quad\Longleftrightarrow\quad
|a|\le\sqrt{c/\lambda_k}.
\]
Thus, the ellipsoid has semiaxis length $\sqrt{c/\lambda_k}$
along $e_k$. A smaller $\lambda_k$ permits a larger perturbation
magnitude $|a|$ without exceeding the same response threshold $c$, so the ellipsoid
extends farther along the corresponding eigenvector $e_k$.
The total volume in \eqref{eq:cs-volume} is $\omega_K$ multiplied by the product
 of these semiaxis lengths.

(3)
The same ellipsoid also describes the perturbations $v$ for which
the self-test loss at $\hat{\theta}_T+v$ is at most $c/2$ above its
minimum. 
Indeed, for
$L_T(\theta)
=\tfrac12\theta^\top A_T\theta-b_T^\top\theta$,
the unique minimizer $\hat{\theta}_T=A_T^{-1}b_T$ satisfies
\begin{equation}\label{eq:cs-curvature}
L_T(\hat{\theta}_T+v)-L_T(\hat{\theta}_T)
=\tfrac12v^\top A_Tv.
\end{equation}
The corresponding parameter vectors form the translated ellipsoid
$\widehat\theta+\mathcal U_c(A_T)$.

\end{remark}

Together with Proposition~\ref{prop:unique}, this clarifies how control
can improve parameter identifiability: 
by changing the visited particle ensembles, control can
provide complementary excitation across time windows to eliminate
common null directions, thereby making the aggregate Gram matrix positive
definite and ensuring uniqueness; once this condition is met,
maximizing $\log\det(A_T)$ minimizes the total volume of weakly
distinguishable parameter perturbations.

\subsection{Estimation error and sampling effects}\label{sec:error}
The preceding subsections characterize uniqueness and parameter distinguishability through the Gram matrix. We now study how stochastic noise and discrete observation affect estimation accuracy. For a single sample ($M=1$), denote the discrete estimator by $\widehat\theta=\widehat\theta_{1,\Delta t}$ and the corresponding continuous-observation quantities by $A_T,b_T$.

Define the stochastic residual
$ 
Z_T=\frac1{NT}\sum_i\int_0^T
F_i(\mathbf X_t)^\top\Sigma\dd W_t^i.
$
For the discrete estimator~\eqref{eq:discrete}, let $E_A=A_{1,\Delta t}-A_T$ and $E_b=b_{1,\Delta t}-b_T$ measure the changes caused by replacing time integrals with observation-grid sums. These changes enter the estimation error through the combined quadrature residual $R_{T,\Delta t}=E_b-E_A\theta^\star$.

\begin{proposition}[Estimation error bound]\label{prop:discrete-error-bound}
Assume that $g_i^\star=F_i\theta^\star$, and that $A_{1,\Delta t}\succ0$. Then the estimator $\widehat\theta=A_{1,\Delta t}^{-1}b_{1,\Delta t}$ satisfies
\begin{equation}\label{eq:exact-snapshot-bound}
\norm{\widehat\theta-\theta^\star}
\le\frac{\norm{Z_T}+\norm{R_{T,\Delta t}}}
{\lambda_{\min}(A_{1,\Delta t})}.
\end{equation}
Here $\norm{\cdot}$ denotes the Euclidean vector norm.
\end{proposition}
\begin{proof}
Start from the integrated cross-term identity~\eqref{eq:integrated-cross}, without taking expectations. For any fixed candidate parameter $\theta$, substituting $g_i^\theta=F_i\theta$ and $g_i^\star=F_i\theta^\star$ makes its left-hand side $\theta^\top A_T\theta^\star$. By~\eqref{eq:energy-vectors} and~\eqref{eq:normal-components}, the endpoint, known-drift, and diffusion terms on its right-hand side together equal $\theta^\top b_T$. Also,
$ 
\frac{M_T^\theta}{T}
=\frac1{NT}\sum_i\int_0^T
(F_i\theta)^\top\Sigma\dd W_t^i
 =\theta^\top Z_T.
$
Thus the integrated cross-term identity becomes
\[
\theta^\top A_T\theta^\star
=\theta^\top b_T+\theta^\top Z_T.
\]
Taking $\theta$ to be each coordinate unit vector gives the vector identity
\begin{equation}\label{eq:exact-residual}
b_T=A_T\theta^\star-Z_T.
\end{equation}
The stochastic term remains because we work with an individual sample rather than its expectation.

Next, subtracting $A_{1,\Delta t}\theta^\star$ from the discrete normal equations and using~\eqref{eq:exact-residual} gives
\[
\begin{aligned}
A_{1,\Delta t}(\widehat\theta-\theta^\star)
&=b_{1,\Delta t}-A_{1,\Delta t}\theta^\star\\
&=(b_T+E_b)-(A_T+E_A)\theta^\star\\
&=-Z_T+E_b-E_A\theta^\star\\
&=-Z_T+R_{T,\Delta t}.
\end{aligned}
\]
Positive definiteness makes $A_{1,\Delta t}$ invertible, so
\begin{equation}\label{eq:exact-snapshot-error}
\widehat\theta-\theta^\star
=A_{1,\Delta t}^{-1}(-Z_T+R_{T,\Delta t}).
\end{equation}
For a symmetric positive-definite matrix, the eigenvalues of its inverse are the reciprocals of its eigenvalues. Hence
$ 
\norm{A_{1,\Delta t}^{-1}}_{\rm op}
=\frac{1}{\lambda_{\min}(A_{1,\Delta t})} 
$,
where $\norm{\cdot}_{\rm op}$ is the operator norm induced by the Euclidean norm. Applying  this operator-norm inequality 
and
the triangle inequality
    to \eqref{eq:exact-snapshot-error} yields
\[
\begin{aligned}
\norm{\widehat\theta-\theta^\star}
&\le\norm{A_{1,\Delta t}^{-1}}_{\rm op}
\norm{-Z_T+R_{T,\Delta t}} \le\frac{\norm{Z_T}+\norm{R_{T,\Delta t}}}
{\lambda_{\min}(A_{1,\Delta t})},
\end{aligned}
\]
which proves~\eqref{eq:exact-snapshot-bound}.
\end{proof}



Proposition~\ref{prop:discrete-error-bound} identifies two residuals
that contribute to estimation error: $Z_T$, which captures accumulated
Brownian noise, and $R_{T,\Delta t}$, which captures the error from
approximating time integrals with discrete observations. Solving the normal equations multiplies the combined residual
$-Z_T+R_{T,\Delta t}$ by $A_{1,\Delta t}^{-1}$, whose operator norm is
$1/\lambda_{\min}(A_{1,\Delta t})$. Thus, for fixed residual norms,
a larger smallest eigenvalue gives a smaller upper bound on the
parameter estimation error.

Because the input can change the visited particle ensembles and
hence the Gram matrix, the bound motivates designing inputs that
strengthen excitation in weakly observed parameter directions.
For given residual magnitudes, an increase in the smallest Gram
matrix eigenvalue reduces residual amplification and tightens
the estimation error bound. Together with the parameter-volume
interpretation in Section~\ref{sec:volume}, this motivates the
log-determinant objective as an observable criterion for
information-directed excitation.

The reward of Section~\ref{sec:cs-ppo} uses the log determinant~\eqref{eq:cs-score}
rather than $\lambda_{\min}$   for two reasons. 
First, the two are
linked: writing $\lambda_{\max}(H) =\lambda_1\ge\dots\ge\lambda_K = \lambda_{\min}(H) \ge0$ for the sorted eigenvalues of a
symmetric positive-semidefinite $H$, then
$$ 
\begin{aligned}
& 
K\,J(H)=\sum_{k=1}^{K}\log(\lambda_k+\varepsilon_G)
\le\log(\lambda_K+\varepsilon_G)+(K-1)\log(\lambda_1+\varepsilon_G)
\\
& 
\Rightarrow ~ 
\lambda_{\min}(H)+\varepsilon_G\ \ge\ \frac{e^{KJ(H)}}{(\lambda_{\max}(H)+\varepsilon_G)^{K-1}} .
\end{aligned}
$$
For a given largest eigenvalue, increasing $J$ therefore raises a guaranteed
lower bound on $\lambda_{\min}$ and, through~\eqref{eq:exact-snapshot-bound}, lowers
the estimation error bound. 
Second, $J$ is better suited as a reward than $\lambda_{\min}$. With
$\lambda_{\min}$ the reward increment would be
$\lambda_{\min}(H_{\ell+1})-\lambda_{\min}(H_\ell)$, which is zero whenever a
step adds information in any direction other than the currently weakest one,
so the learning signal is sparse; moreover, $\lambda_{\min}(H)$ is not
differentiable at matrices whose smallest eigenvalue has multiplicity greater
than one. In contrast, $J$ is smooth on the positive-semidefinite cone and
every direction contributes, with weight
$\partial J/\partial\lambda_k=\tfrac1K(\lambda_k+\varepsilon_G)^{-1}$: the
same increase of an eigenvalue is rewarded more when that eigenvalue is small,
so the weakest direction receives the largest weight, and the weight of a
direction that has not yet been excited is capped at $ \frac{1}{K\varepsilon_G}$.

\section{Numerical experiments}\label{sec:experiments}

 This section presents simulation examples to validate our method.
Two systems with four unknown potential coefficients are considered: a one-dimensional interacting diffusion with two localized side wells, and a planar flock whose acceleration depends on anticipated positions. In both cases, the true potentials are representable using the prescribed basis functions.
The results evaluate coefficient estimation provided by different control inputs.

\subsection{Experimental setup}\label{sec:exp-systems}

\paragraph{System 1: One-dimensional interacting diffusion}
\label{sec:exp-scalar}
For $X^i,u^i\in\R$, define $g_\pm(x)=\exp[-(x\mp c)^2/(2w^2)]$, with known $c=3$ and $w=0.35$. The potentials and true coefficients are
\begin{equation}\label{eq:truth}
\begin{aligned}
V_\theta(x)&=\theta_1x^2/2+\theta_2g_-(x)+\theta_3g_+(x),\\
\Phi_\theta(r)&=\theta_4r^2/2,
\\
\theta&=(\theta_1,\theta_2,\theta_3,\theta_4)^\top,\quad
\theta^\star =(0.5,-2,-2.5,0.6)^\top.
\end{aligned}
\end{equation}
With these potentials, the controlled dynamics are
\begin{equation}\label{eq:exp-scalar-sde}
\dd X^i=\left[-\nabla V_{\theta^\star}(X^i)
-\frac1N\sum_{j\ne i}\nabla\Phi_{\theta^\star}(X^i-X^j)
+u^i\right]\dd t+\sigma\dd W^i.
\end{equation}
$V_{\theta^\star}$ has three wells, near $0$ and $\pm c$, while $\Phi_{\theta^\star}$ attracts particles toward their mean. A cloud remaining near zero supplies little information about the two side-well coefficients, motivating excitation toward both sides. This is \eqref{eq:sde} with $d=d_u=1$, $A_0=0$, $D=B=1$, and $\Sigma=\sigma$.

\paragraph{System 2: Anticipated-position flock}
\label{sec:exp-flock}
Each observed state is $X^i=((Z^i)^\top,(V^i)^\top)^\top\in\R^4$, with planar position $Z^i$ and velocity $V^i$. Following the anticipated-position mechanism of
\cite{shu},
define $q^i=Z^i+\tau V^i$ and $s^{ij}=q^i-q^j$, with known $\tau=0.5$. No future observation is required. 
With known $\ell_{\rm int}=1$,  
the potentials and true coefficients are
\begin{equation}\label{eq:exp-flock-potentials}
\begin{aligned}
U_\theta(q)&=\kappa_2\norm q^2/2+\kappa_4\norm q^4/4,\\
\Psi_\theta(s)&=a\norm s^2/2+b\ell_{\rm int}^2e^{-\norm s^2/(2\ell_{\rm int}^2)},\\
\theta&=(\kappa_2,\kappa_4,a,b)^\top,\quad
\theta^\star=(0.5,0.4,1,2)^\top.
\end{aligned}
\end{equation}
With these potentials, the controlled dynamics are
\begin{equation}\label{eq:exp-flock-sde}
\begin{aligned}
\dd Z^i&=V^i\dd t,\\
\dd V^i&=\left[-\nabla U_{\theta^\star}(q^i)
-\frac1N\sum_{j\ne i}\nabla\Psi_{\theta^\star}(s^{ij})+u^i\right]\dd t
+\sigma\dd W^i,
\end{aligned}
\end{equation}
where $u^i,W^i\in\R^2$. The pair force is repulsive at short anticipated separations and attractive at long ones. Small separations make the attraction and repulsion features nearly collinear, motivating exploration across separation scales. 
This system is an instance of~\eqref{eq:sde} with $d=4$, $d_u=d_W=2$, and
\begin{equation}\label{eq:flock-matrices}
\begin{aligned}
&A_0=\begin{bmatrix}0&I_2\\0&0\end{bmatrix},\quad
 B=\begin{bmatrix}0\\ I_2\end{bmatrix},\\
&\Sigma=\begin{bmatrix}0\\ \sigma I_2\end{bmatrix},\quad
 D=\begin{bmatrix}\tau I_2&-I_2\\-I_2&2\tau^{-1}I_2\end{bmatrix}\succ0
 \ \text{(Schur complement $\tau^{-1}I_2$)}.
\end{aligned}
\end{equation}
Here $V^\star(X^i)=U_{\theta^\star}(q^i)$ and
$\Phi^\star(X^i-X^j)=\Psi_{\theta^\star}(s^{ij})$, where
$s^{ij}=(Z^i-Z^j)+\tau(V^i-V^j)$ is a function of $X^i-X^j$ as required
in~\eqref{eq:force}. Since both potentials depend on $X^i=(Z^i,V^i)$ only
through $q^i=Z^i+\tau V^i$, the gradient force in~\eqref{eq:force} is
$g_i^\star=
\begin{bmatrix}f_i\\ \tau f_i\end{bmatrix}
$ 
with
$f_i=\nabla U_{\theta^\star}(q^i)+\frac1N\sum_{j\ne i}\nabla\Psi_{\theta^\star}(s^{ij})$,
and~\eqref{eq:flock-matrices} gives $A_0X^i=
\begin{bmatrix} V^i \\  0 \end{bmatrix}
$, $Dg_i^\star=
\begin{bmatrix} 0\\   f_i\end{bmatrix}
$ ,
$Bu^i=
\begin{bmatrix} 0 \\ u^i\end{bmatrix}
$ 
, so~\eqref{eq:sde} reduces to~\eqref{eq:exp-flock-sde}; the
diffusion $C=\operatorname{diag}(0,\sigma^2I_2)$ is degenerate.
Any symmetric $D$ with $D
\begin{bmatrix}  I_2 \\ \tau I_2\end{bmatrix}
=
\begin{bmatrix}  0 \\  I_2\end{bmatrix}
$ 
serves equally and yields
the same loss, because $D$ enters~\eqref{eq:risk}--\eqref{eq:gram} only through
$[I_2,\ \tau I_2]\,D\,
\begin{bmatrix}  I_2 \\ \tau I_2\end{bmatrix}
=\tau I_2$.
The anticipation parameter $\tau>0$ is what places this second-order system
within the mobility structure of~\eqref{eq:sde}: it couples the force to both
position and velocity, which is exactly what a symmetric positive-definite $D$
requires. Position-only forces ($\tau=0$) admit no such $D$ and lie outside
the present framework.

Both systems satisfy the assumptions of Proposition~\ref{prop:risk}.
The inputs are bounded by $u_{\max}$ and are functions of the current
observations. In System~1 the force is linear in $x$ up to bounded Gaussian
terms, hence globally Lipschitz, so the solution has finite moments on
$[0,T]$. In System~2 the potential grows like $\|q\|^4$ at infinity because
$\kappa_4>0$, so the particles cannot escape and the solution again has
finite moments on $[0,T]$. Every candidate force is a polynomial or a bounded
function of the state, so the integrability condition in
Proposition~\ref{prop:risk} holds.

\paragraph{Policy training}
Training rollouts simulate the systems with coefficients sampled
independently per rollout as
\begin{equation}\label{eq:train-prior}
\begin{aligned}
&
\tilde\theta_k=\theta^\star_k(1+s_ka_k),
\\
&  a_k\sim U[0.2,0.4], 
P(s_k=+1)=\tfrac23,\quad P(s_k=-1)=\tfrac13 .
\end{aligned}
\end{equation}
Each coefficient keeps its sign, and its magnitude is 60--80\% of the true
value with probability $\tfrac13$ and 120--140\% with probability $\tfrac23$; the training mean is therefore biased to $1.1\,\theta^\star$, and the true
coefficients are never simulated.
Validation rollouts are also sampled  from~\eqref{eq:train-prior}. 
Test
samples are generated from the true unknown system with $\theta^\star$, with the same
initial conditions and noise as for the baselines.

\paragraph{Evaluation protocol and metrics}
\label{sec:exp-protocol}
Table~\ref{tab:exp-settings} summarizes the numerical settings. Each independent sample starts from $X_{t_0}^{i,m}=\eta^m+\xi^{i,m}$, where $\eta^m\sim\mathcal N(0,0.25^2I_d)$ and $\xi^{i,m}\overset{\rm iid}{\sim}\mathcal N(0,0.1^2I_d)$ are independent, with $d=1$ or $4$.   
Euler--Maruyama simulation uses a finer step than observation and input holding. 
Particle indices are unpaired across observation times; flock position and velocity are observed jointly for each particle at each time.
Commands act on their current recipients for one interval without supplying cross-time identities.

\begin{table}[!htbp]
\centering\small
\begin{tabularx}{\linewidth}{@{}>{\raggedright\arraybackslash}p{0.45\linewidth}X@{}}
\toprule
Setting & Value\\
\midrule
Particles and horizon & $N=10$, $T=1$\\
Noise amplitude & $\sigma=0.4$\\
Simulation / observation step & $10^{-4}$ / $\Delta t=10^{-3}$\\
Control intervals & $L=T/\Delta t=1000$ ($L+1=1001$ observation times) \\
Training coefficients & \eqref{eq:train-prior} \\
Input bound & $u_{\max}=10$ per coordinate\\
Scalar sample counts $M$ & $50,100,1000,2000$\\
Flock sample counts $M$ & $1,50,100,500,$ $1000,2000$\\
Scalar training / validation & 1600 / 110 sample evaluations\\
Flock training / validation & 1600 / 110 sample evaluations\\
Reward parameters & $\varepsilon_G=10^{-4}$, $\beta_u=0.01$\\
\bottomrule
\end{tabularx}
\caption{Evaluation settings and PPO construction budgets. Validation counts are sample evaluations over ten reused seeds.}\label{tab:exp-settings}
\end{table}

We compare four input strategies: Passive ($u^i=0$), Random (independent uniform commands in $[-10,10]$ per coordinate, particle, sample, and interval), Sinusoid, and PPO with a frozen policy. The sinusoidal input is $10\sin(2\pi t_\ell)$ for the scalar system and $10(\sin(2\pi t_\ell),\cos(2\pi t_\ell))^\top$ for the flock ---
this amplitude equals the common bound $u_{\max}$ and one period spans the
horizon, which gives the largest excursion compatible with a return toward
the initial region; Section~\ref{sec:exp-information} shows that no open-loop
sinusoid within the bound can visit both wells of System~1.
All active methods have the same componentwise bound. Reported input energy is the mean $M^{-1}\sum_{m=1}^M E_u^m$ of the per-sample quantity in~\eqref{eq:cs-telescope}.


The evaluation metrics include relative coefficient error to assess
identification accuracy, the unregularized minimum eigenvalue
 $ 
\lambda_M=\lambda_{\min}(A_{M,\Delta t})$,
and the
regularized log determinant 
$\mathcal D_M=\log\det(A_{M,\Delta t}+\varepsilon_GI_K)=K\,J(A_{M,\Delta t})$
of the pooled Gram matrix
to assess information in the weakest direction and across all directions respectively, and input energy to measure control effort.

\subsection{Results and analysis}\label{sec:exp-results}

\subsubsection{Identification accuracy and force reconstruction}
\label{sec:exp-accuracy}\label{sec:exp-scalar-results}\label{sec:exp-flock-results}

\begin{figure}[!htbp]
\centering\includegraphics[width=\columnwidth]{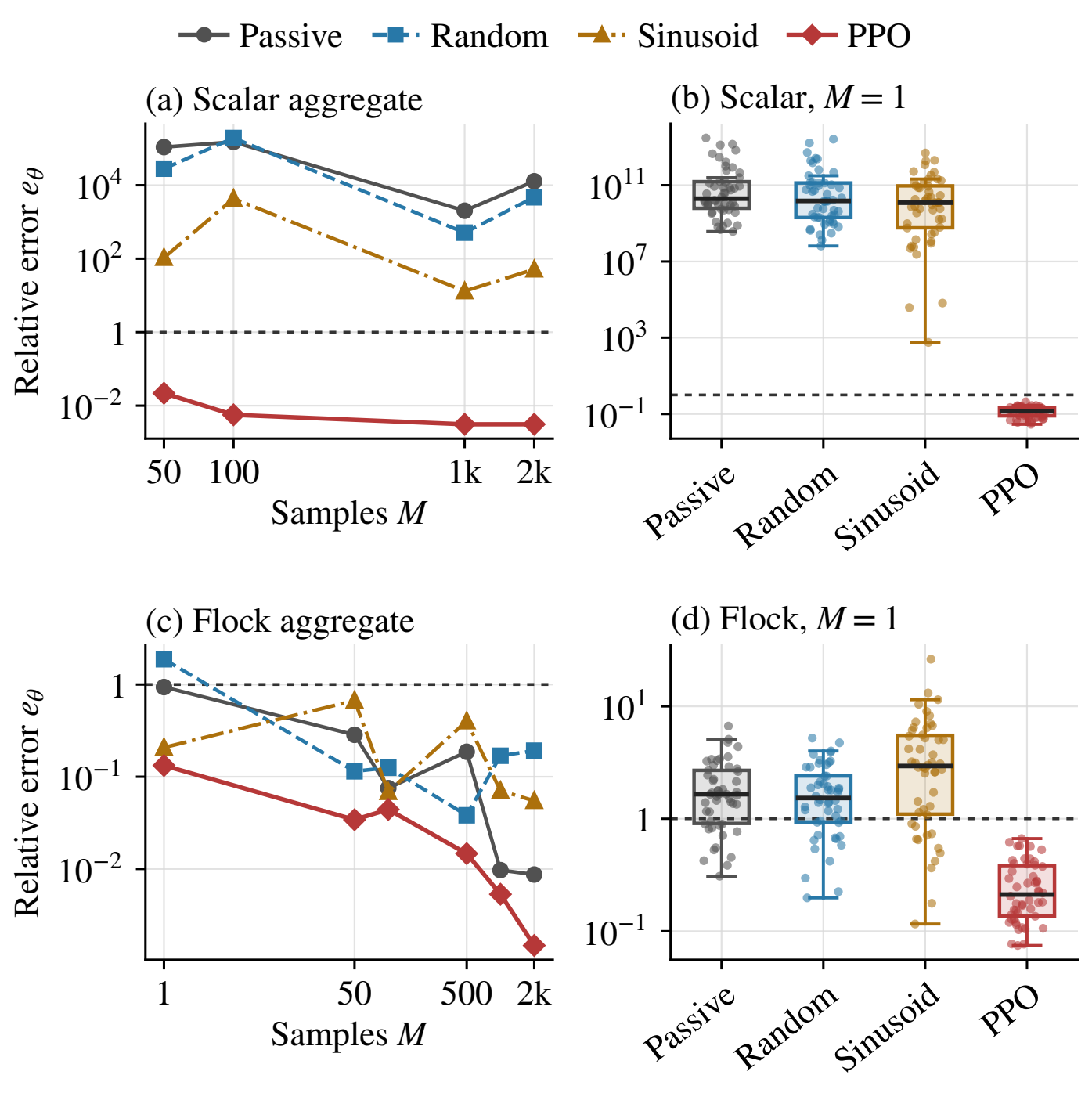}
\caption{Relative coefficient identification error for the scalar system (top) and flock (bottom), with $N=10$. 
Panels (a,c) show 
one aggregate fit for every tested sample count $M$
; (b,d) show 50 independent fits for single test sample $M=1$. 
Boxes show medians and quartiles, with 1.5-IQR whiskers and all observations overlaid. Error axes are logarithmic; dashed lines mark $100\%$ error.}
\label{fig:accuracy-overview}\label{fig:errors}\label{fig:flock-error}\label{fig:scalar-single}\label{fig:flock-single}
\end{figure}

We assess both aggregate identification accuracy and sample-to-sample variability using the relative coefficient error
$ 
e_\theta(M)
=
\frac{\norm{\widehat\theta_{M,\Delta t}-\theta^\star}}
     {\norm{\theta^\star}}$.
Figure~\ref{fig:accuracy-overview}(a,c) compares this error across independent-sample counts $M$, with each aggregate estimate obtained by averaging the sample normal equations before inversion. Figure~\ref{fig:accuracy-overview}(b,d) shows the distributions of 50 independent $M=1$ estimates. 
For each system, the policy is trained using PPO on $1600$ training samples and
selected through $110$ validation sample evaluations, all with coefficients
sampled from~\eqref{eq:train-prior},
and then policy parameters are frozen across all
test sample sizes and single-sample trials.

\begin{figure}[!htbp]
\centering\includegraphics[width=\columnwidth]{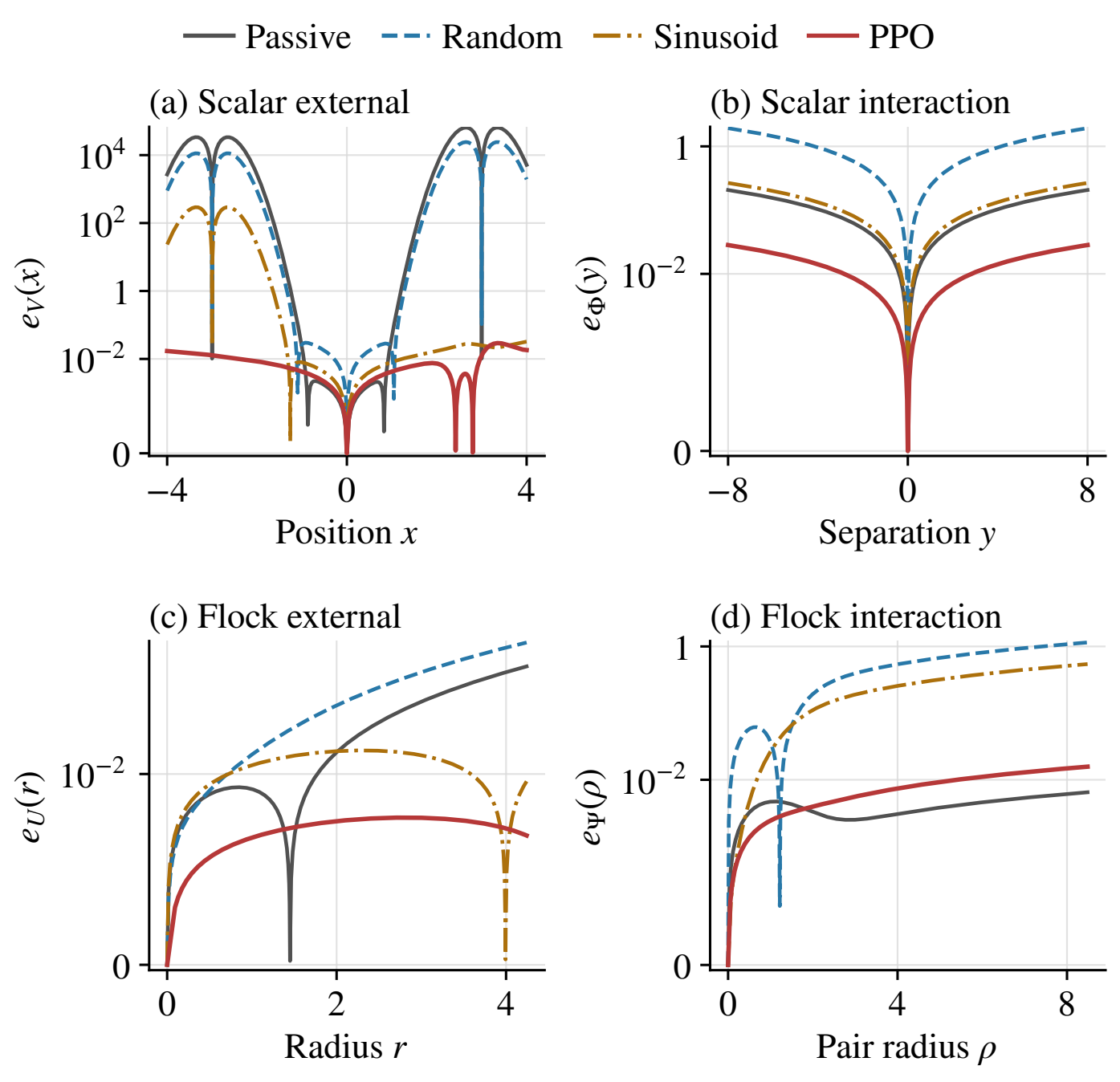}
\caption{Pointwise potential-force errors from the $M=2000$ estimates in Table~\ref{tab:comparison}: external (left) and interaction (right) errors for the scalar system (top) and flock (bottom). Flock profiles show   radial $D$-norm errors. Vertical axes are   symmetric logarithmic, with a linear region below $10^{-4}$.}
\label{fig:force-overview}\label{fig:scalar-force-error}\label{fig:flock-external-force-error}\label{fig:flock-interaction-force-error}
\end{figure}

The results demonstrate the benefits of learned excitation for coefficient recovery and show that the error decreases further as more independent samples are pooled.
In the scalar system, learned excitation provides a clear identification advantage over the tested baselines, already evident at $M=1$ and maintained across all tested aggregate sample counts (Figure~\ref{fig:accuracy-overview}(a,b)). For the single-sample fits, PPO achieves a median relative error of $0.1401$, compared with baseline medians of order $10^{10}$. With increasing $M$, PPO continues to give the smallest aggregate error, which decreases overall from $2.179\%$ at $M=50$ to $0.311\%$ at $M=2000$. The flock results show a substantial benefit at every sample count (Figure~\ref{fig:accuracy-overview}(c)): at $M=50$, PPO reduces the error from $11.45\%$ for the best baseline, Random, to $3.4\%$, and it gives the smallest error at every tested $M$, including the single realization at $M=1$ ($13\%$ versus $20.73\%$ for Sinusoid, the best baseline there). Larger $M$ also permits accurate passive estimation, reaching $0.870\%$ at $M=2000$, but PPO reaches $0.148\%$. PPO itself exhibits an overall downward error trend as $M$ increases, from $13\%$ at $M=1$ to $0.148\%$ at $M=2000$.
Detailed 
coefficient estimates 
are reported in Table~\ref{tab:comparison}.

The single-sample results further show that PPO improves the consistency of recovery across the tested realizations (Figure~\ref{fig:accuracy-overview}(b,d)). In the scalar system, the PPO errors have an interquartile range of $[0.0800,0.2147]$, compared with the much broader Passive interval $[6.143\times10^9,1.531\times10^{11}]$. PPO also outperforms each baseline in all 50 paired scalar trials. In the flock, PPO lowers the median error from $165\%$ under Passive to about $21\%$ and is the only method whose median error lies below $100\%$.

The force profiles show that PPO substantially improves recovery of the localized scalar forces, and its gains in the flock also depend on the force component. 
Using the $M=2000$ samples in self-test loss, PPO reduces the maximum plotted scalar external-force error from $6.482\times10^4$ under Passive to $2.928\times10^{-2}$, with the largest baseline discrepancies occurring near the side wells (Figure~\ref{fig:force-overview}(a)). In the flock, 
PPO reduces the maximum external-force error by more than two orders of magnitude relative to Passive, while the maximum interaction-force error under PPO is approximately $2.4$ times that under Passive (Figure~\ref{fig:force-overview}(c,d)).

\subsubsection{State exploration, information accumulation, and input energy}\label{sec:exp-information}





\begin{figure}[!htb]
\centering\includegraphics[width=\linewidth]{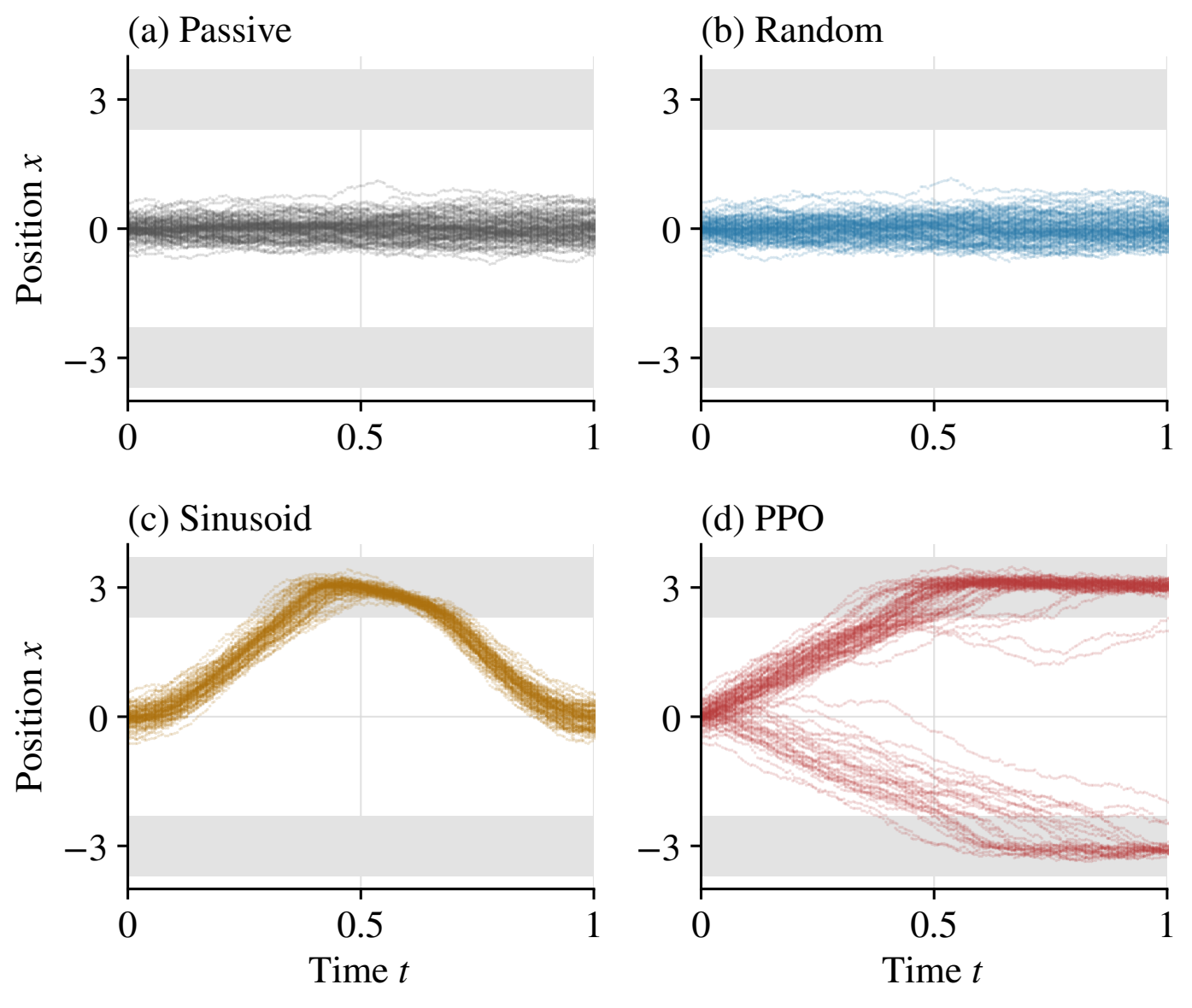}
\caption{
Scalar particle positions from ten randomly selected samples, each containing $N=10$ particles.
Gray bands mark the side-well neighborhoods $|x\mp3|\le0.7$.}
\label{fig:scalar-states}
\end{figure}

\paragraph{State coverage}
Figure~\ref{fig:scalar-states} shows particle positions over time in the scalar system under the four input strategies. Each panel overlays ten samples, with $N=10$ particles per sample, to make the state coverage visible.

Under PPO, particles reach both side-well neighborhoods in all ten displayed samples: PPO positions range from $-3.364$ to $3.489$, with particles entering both the left well near $x=-3$ and the right well near $x=3$.  
By contrast, none of the three baselines reaches both neighborhoods in any of these samples. 
Passive and Random reach neither neighborhood; Sinusoid reaches the right neighborhood in all ten samples and the left in none. 
These failures are structural. 
Random inputs are zero-mean
and uncorrelated across intervals, so their net displacement is negligible.
The sinusoid is the same for all particles, so the ensemble moves as one block
(Figure~\ref{fig:scalar-states}(c)) and can reach the two wells only in turn;
a half period of $u_{\max}\sin(\omega t)$ moves the block by at most about
$2u_{\max}/\omega$, not enough to travel from the well at $x=3$ to the one at
$x=-3$ within the horizon. 
The feedback policy instead pushes each particle
toward its nearer well (Figure~\ref{fig:scalar-states}(d)), so the population
splits and both wells are covered with about $60\%$ of the input energy of the sinusoid
(Table~\ref{tab:flock-behavior}). 
The coverage difference shows directly in the
side-well coefficients at $M=2000$: Sinusoid estimates the right-well
coefficient accurately as $-2.5034$ (true value $-2.5$) but the left-well
coefficient as $167.62$ (true value $-2$), whereas PPO gives $-1.9999$ and
$-2.4914$ (Table~\ref{tab:comparison}).

\begin{table}[!t]
\centering\small\setlength{\tabcolsep}{3pt}
\caption{Estimated coefficients at $M=2000$, with ground truth for comparison. Columns follow $(\theta_1,\theta_2,\theta_3,\theta_4)$ for the scalar system and $(\kappa_2,\kappa_4,a,b)$ for the flock.}
\label{tab:comparison}\label{tab:scalar-coefficients}\label{tab:flock-coefficients}
\resizebox{\columnwidth}{!}{%
\begin{tabular}{llrrrr}
\toprule
System & Input & $\widehat\theta_1$ & $\widehat\theta_2$ & $\widehat\theta_3$ & $\widehat\theta_4$ \\
\midrule
Scalar & Passive & $0.49659$ & $1.9540\times10^{4}$ & $3.7401\times10^{4}$ & $0.62584$ \\
 & Random & $0.46471$ & $6.5555\times10^{3}$ & $1.4014\times10^{4}$ & $0.35897$ \\
 & Sinusoid & $0.49182$ & $167.62$ & $-2.5034$ & $0.63333$ \\
 & PPO & $0.49573$ & $-1.9999$ & $-2.4914$ & $0.60354$ \\
 & Truth & $0.5$ & $-2$ & $-2.5$ & $0.6$ \\
\midrule
Flock & Passive & $0.48373$ & $0.40773$ & $1.0011$ & $1.9909$ \\
 & Random & $0.48738$ & $0.38549$ & $1.1912$ & $2.402$ \\
 & Sinusoid & $0.47971$ & $0.40127$ & $1.0905$ & $2.0874$ \\
 & PPO & $0.49825$ & $0.40007$ & $0.99738$ & $2.0014$ \\
 & Truth & $0.5$ & $0.4$ & $1$ & $2$ \\
\bottomrule
\end{tabular}
 
}
\end{table}

\begin{figure}[!htbp]
\centering\includegraphics[width=\columnwidth]{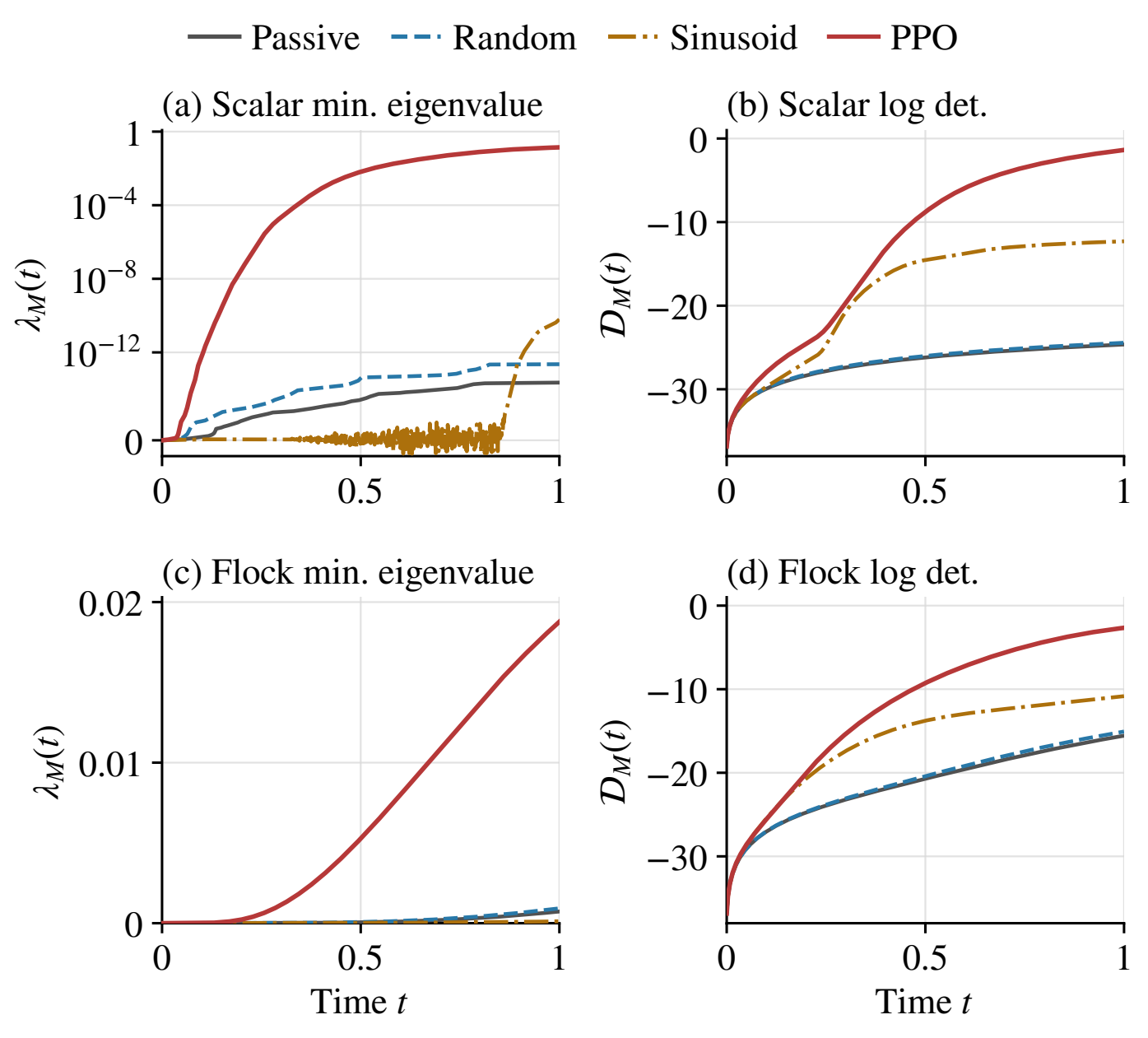}
\caption{Information accumulation over $M=2000$ test samples in the scalar system (a,b) and flock (c,d): minimum unregularized eigenvalue $\lambda_M(t)$ (left) and regularized log determinant $\mathcal D_M(t)$ (right), with $\varepsilon_G=10^{-4}$. Panel (a) uses a symmetric logarithmic scale, linear for $|\lambda_M(t)|\le10^{-16}$.}
\label{fig:information-overview}\label{fig:information}\label{fig:flock-information}
\end{figure}

\paragraph{Information gain}
In the overdamped scalar system, PPO gives the largest minimum   eigenvalue and the largest   log determinant among the four methods (Figure~\ref{fig:information-overview}(a,b)). 
With $M=2000$ independent test samples, its   minimum eigenvalue is $1.396\times10^{-1}$, compared with $6.266\times10^{-11}$ for Sinusoid, the strongest baseline on this measure: an increase of approximately $2.2\times10^9$. PPO's   log determinant is $-1.370$, compared with $-12.298$ for Sinusoid, which also has the largest log determinant among the baselines (Table~\ref{tab:m2000-information}).

PPO also gives the largest values of both information measures in the flock system (Figure~\ref{fig:information-overview}(c,d)). Its   minimum eigenvalue is $1.877\times10^{-2}$, approximately $25.7$ times the Passive value $7.292\times10^{-4}$ and $154$ times the Sinusoid value $1.216\times10^{-4}$. Its   log determinant is $-2.656$, compared with $-10.833$ for Sinusoid, the largest baseline value. These results show that PPO improves both information in the weakest coefficient direction and the combined information score across all four directions.

\begin{table}[!htb]
\centering\small\setlength{\tabcolsep}{2.5pt}
\caption{Unregularized minimum eigenvalue, regularized log determinant ($\varepsilon_G=10^{-4}$), and mean input energy for $M=2000$ test samples per system.}
\label{tab:m2000-information}\label{tab:flock-behavior}
\begin{tabular}{@{}llrrr@{}}
\toprule
System & Input & $\lambda_{\min}(A_{M,\Delta t})$ & $\mathcal D_M(T)$ & $E_u$\\
\midrule
Scalar & Passive & $2.282\times10^{-14}$ & -24.643 & 0.00 \\
 & Random & $2.272\times10^{-13}$ & -24.439 & 33.33 \\
 & Sinusoid & $6.266\times10^{-11}$ & -12.298 & 50.00 \\
 & PPO & $1.396\times10^{-1}$ & -1.370 & 29.33 \\
\midrule
Flock & Passive & $7.292\times10^{-4}$ & -15.563 & 0.00 \\
 & Random & $9.253\times10^{-4}$ & -15.067 & 66.67 \\
 & Sinusoid & $1.216\times10^{-4}$ & -10.833 & 100.00 \\
 & PPO & $1.877\times10^{-2}$ & -2.656 & 34.06 \\
\bottomrule
\end{tabular}

\end{table}

\paragraph{Input energy}
Despite achieving greater information gain, PPO consumes less input energy than Random and Sinusoid in both systems.
For the $M=2000$ tests in Table~\ref{tab:m2000-information}, 
scalar PPO uses $E_u=29.33$, compared with $33.33$ for Random and $50.00$ for Sinusoid. 
Flock PPO uses $34.06$, compared with $66.67$ for Random and $100.00$ for Sinusoid. 
Relative to Sinusoid, these values represent energy reductions of $41.3\%$ in the scalar system and $65.9\%$ in the flock system. 
PPO therefore obtains higher values of both information measures while using less input energy than these two active baselines.

\subsubsection{Sensitivity to particle and sample counts}\label{sec:exp-sensitivity}
We assess sensitivity to the particle count $N\in\{10,20,50,100\}$ and independent test-sample count $M\in\{50,100,1000,2000\}$ in the scalar system of~\eqref{eq:exp-scalar-sde}, keeping the remaining physical and observation settings fixed (
$\Delta t = 10^{-3}$, 
$L=1000$ control intervals). A separate PPO policy is trained with $160N$ samples for each $N$, with coefficients sampled from~\eqref{eq:train-prior}, and then frozen for all test counts $M$.  

\begin{figure}[!htbp]
\centering\includegraphics[width=\columnwidth]{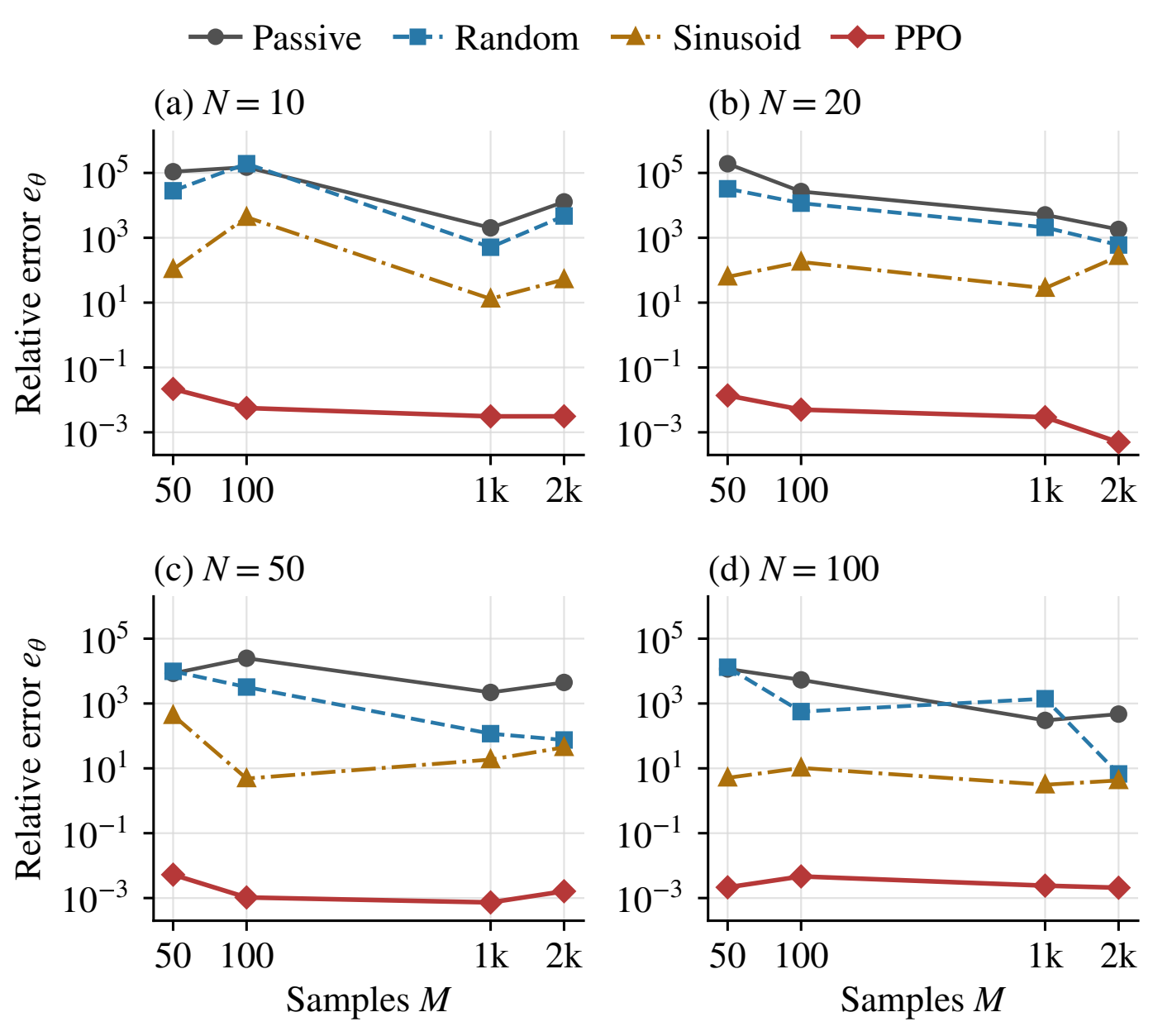}
\caption{Relative coefficient error across different sample counts $M$ for (a) $N=10$, (b) $N=20$, (c) $N=50$, and (d) $N=100$. 
Each point is one aggregate fit. PPO is trained separately at each $N$ using $160N$ samples. Both axes are logarithmic.}
\label{fig:particle-errors}\label{fig:particle-errors-small}\label{fig:particle-errors-large}
\end{figure}

PPO retains its accuracy advantage across all tested particle and independent-sample counts. Figure~\ref{fig:particle-errors} shows that PPO has the lowest realized coefficient error in all 16 configurations, with errors ranging from $0.0495\%$ to $2.179\%$. Even at $N=100,M=2000$, the best baseline, Sinusoid, has $423.7\%$ error, compared with $0.208\%$ for PPO.

Larger particle populations $N$ improve coefficient accuracy under PPO control and allow errors below $1\%$ with fewer independent samples $M$. 
At every tested $M$, the error at $N=100$ is lower than at $N=10$. 
This advantage is useful when independent samples are limited, since larger $N$ improves sample efficiency: at $M=50$, the errors for $N=50$ and $100$ are $0.526\%$ and $0.214\%$, respectively, whereas for $N=10$ and $20$, $M=100$ is the smallest tested sample count that yields an error below $1\%$.



At fixed $N$, pooling more samples lowers the coefficient error under PPO control most where it starts highest. Raising $M$ from $50$ to $2000$ reduces the error from $2.179\%$ to $0.311\%$ at $N=10$ and from $1.370\%$ to $0.0495\%$ at $N=20$. At $N=100$, the error with only $50$ samples ($0.214\%$) is already below what $N=10$ reaches with $2000$ ($0.311\%$), and further samples change it little ($0.208\%$ at $M=2000$): a larger population attains with $50$ samples the accuracy that a small population needs $2000$ for.
Both the $N$ and $M$ comparisons describe overall trends, with some variation between individual fits.

\begin{table}[!htbp]
\centering\small\setlength{\tabcolsep}{3pt}
\caption{Single-sample estimation error under PPO over 50 independent trials per particle count $N$. Errors are dimensionless; Q25--Q75 gives the quartiles. The last two columns count errors below $10\%$ and $100\%$.}
\label{tab:particle-single}
\begin{tabular}{@{}rrrrr@{}}
\toprule
$N$ & Median & Q25--Q75 & $e_\theta<0.1$ & $e_\theta<1$\\
\midrule
10 & 0.1401 & 0.0800--0.2147 & 21/50 & 50/50 \\
20 & 0.0862 & 0.0604--0.1107 & 36/50 & 50/50 \\
50 & 0.0508 & 0.0347--0.0668 & 48/50 & 50/50 \\
100 & 0.0375 & 0.0271--0.0469 & 50/50 & 50/50 \\
\bottomrule
\end{tabular}

\end{table}

\begin{figure}[!htbp]
\centering\includegraphics[width=0.9\linewidth]{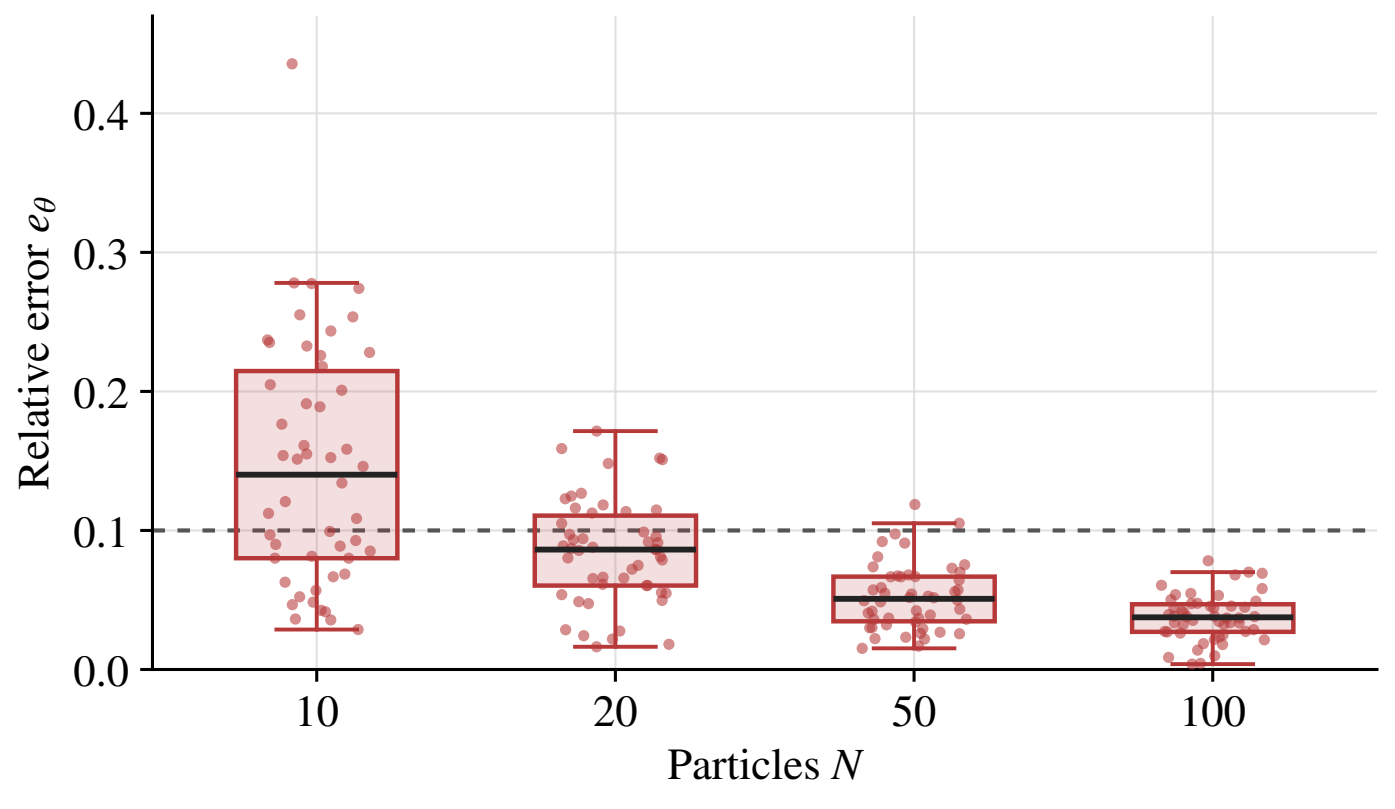}
\caption{Single-sample relative coefficient errors under  PPO for $N=10,20,50,100$, with 50 independent fits per $N$. 
Boxes show medians and quartiles, with 1.5-IQR whiskers and all observations overlaid. The dashed line marks $10\%$ error.}
\label{fig:particle-single-boxscatter}
\end{figure}

For single-sample estimation ($M=1$), increasing the number of particles $N$ from 10 to 100 improves both accuracy and consistency across samples. 
To quantify sample-to-sample variability, we perform 50 independent single-sample fits for each $N$ and summarize the resulting coefficient error distributions using box plots (Table~\ref{tab:particle-single} and Figure~\ref{fig:particle-single-boxscatter}). 
The median coefficient error decreases with $N$, from $14.01\%$ at $N=10$ to $8.62\%$, $5.08\%$, and $3.75\%$ at $N=20$, $50$, and $100$, an overall reduction of about $73\%$ from $N=10$ to $N=100$.
As $N$ increases from 10 to 100, the error distribution becomes narrower, with the interquartile range decreasing by about $85\%$; the fraction of trials with error below $10\%$ also rises from $42\%$ to $100\%$.
Thus, increasing $N$ makes single-sample estimation more accurate and less sensitive to differences between samples.

\section{Conclusion and future work}\label{sec:discussion}
 
In this work, we proposed a framework for potential identification from unlabeled particle evolution data. 
We extended self-test estimation to generalized controlled interacting SDEs
covering overdamped and anticipation-coupled underdamped dynamics, enabling joint recovery of external and interaction potentials without particle matching or numerical time derivatives.
We characterized uniqueness through positive definiteness of the accumulated Gram matrix and quantified parameter distinguishability via the volume of weakly distinguishable parameter perturbations.
Our error analysis also distinguished stochastic fluctuations from observation-grid quadrature error.
Guided by these results, we used proximal policy optimization (PPO) to learn a feedback input policy that steers particles toward informative particle ensembles. The policy uses current observed particle ensembles and accumulated Gram matrix information, with a reward that balances increases in the log determinant against control energy. 
Training uses coefficients sampled from a prior rather than their true
values; once trained, the policy is fixed to collect new data for self-test
estimation.

Our experimental findings suggest that learned excitation can improve potential identification when passive sampling leaves informative regions of the state space poorly explored.
This benefit is evident in the scalar system, where PPO gives the lowest coefficient errors across all tested sample counts and substantially improves reconstruction of both localized side wells.
PPO also substantially improves coefficient recovery in the flock system.
These gains extend to the 50 independent single-sample trials per system, where lower mean and median coefficient errors and narrower interquartile ranges demonstrate more accurate and consistent recovery across realizations.
The state-coverage results help explain the scalar improvement: PPO drives particles toward both side-well neighborhoods, which passive sampling leaves poorly explored.
Consistent with this interpretation, the information diagnostics at $M=2000$ show that PPO yields larger minimum Gram eigenvalues and  log determinants in both systems while using less input energy than the random and sinusoidal controls.
The scalar sensitivity study further shows that PPO retains its accuracy advantage across the tested particle and sample counts.
Under policies trained separately for each particle count $N$, larger $N$ generally improves coefficient accuracy and reduces error variability across single-sample trials.
At fixed $N$, pooling more independent samples $M$ also lowers coefficient error overall.
Together, these findings support learned excitation as a means of collecting more informative particle evolution data and improving the accuracy and consistency of potential identification.





Future work includes learning excitation policies alongside potential representations from data.
Input design can incorporate state and energy constraints.
Another direction is to optimize excitation inputs and observation times to reduce observation-grid quadrature error.
Theoretical questions concern sharper finite-sample error bounds under adaptive excitation and the connection between policy optimization and estimation accuracy.
Potential applications involve adapting the framework to experimental data from single-cell gene-knockout studies~\cite{datlinger2017} and optical-tweezer measurements of interacting colloidal particles~\cite{zhang2024optical}.


\appendix

\bibliographystyle{elsarticle-num}
\bibliography{referencessss}
\end{document}